\documentclass[journal]{IEEEtran}
\usepackage[numbers]{natbib}

\usepackage[ruled,norelsize]{algorithm2e}

\makeatletter
\newcommand{\removelatexerror}{\let\@latex@error\@gobble}
\makeatother

\usepackage{amsmath}
\usepackage{amssymb}
\usepackage{amsfonts}
\usepackage{caption}

\usepackage{mathrsfs}
\usepackage[hidelinks]{hyperref}
\usepackage{subcaption}
\usepackage[table]{xcolor}

\ifCLASSINFOpdf
   \usepackage[pdftex]{graphicx}
\else
\fi
\begin{document}
%
\title{Random Ferns for Semantic Segmentation of PolSAR Images}
%
%
%

\author{Pengchao Wei 
        and
        Ronny H\"ansch,~\IEEEmembership{Senior Member,~IEEE}
\thanks{P. Wei is with the Electronic Engineering department in School of Electronic Information and Electrical Engineering, Shanghai Jiao Tong University (SJTU), China.}
\thanks{R.H\"ansch is with the SAR Technology department of the German Aerospace Center (DLR), Oberpfaffenhofen, Germany.}
\thanks{This is the author's version of the article as accepted for publication. Link to original: \url{https://ieeexplore.ieee.org/document/9627989}}
\thanks{DOI 10.1109/TGRS.2021.3131418}
\thanks{0196-2892 (c) 2021 IEEE. Personal use is permitted, but republication/redistribution requires IEEE permission.}}

%
%

\markboth{IEEE TRANSACTION ON GEOSCIENCE AND REMOTE SENSING}%
{Wei \MakeLowercase{\textit{et al.}}: Random Ferns for Semantic Segmentation of PolSAR Images}
%



\maketitle

\begin{abstract}
Random Ferns - as a less known example of Ensemble Learning - have been successfully applied in many Computer Vision applications ranging from keypoint matching to object detection.  
This paper extends the Random Fern framework to the semantic segmentation of polarimetric synthetic aperture radar images. 
By using internal projections that are defined over the space of Hermitian matrices, the proposed classifier can be directly applied to the polarimetric covariance matrices without the need to explicitly compute predefined image features. 
Furthermore, two distinct optimization strategies are proposed: The first based on pre-selection and grouping of internal binary features before the creation of the classifier; and the second based on iteratively improving the properties of a given Random Fern. 
Both strategies are able to boost the performance by filtering features that are either redundant or have a low information content and by grouping correlated features to best fulfill the independence assumptions made by the Random Fern classifier. 
Experiments show that results can be achieved that are similar to a more complex Random Forest model and competitive to a deep learning baseline.
\end{abstract}
\begin{IEEEkeywords}
Random ferns, ensemble classifier, semantic segmentation, PolSAR image, fern optimization.
\end{IEEEkeywords}

%
\IEEEpeerreviewmaketitle

\section{Introduction\label{sec:section1}}
\IEEEPARstart{R}{emote} sensing and in particular the usage of satellite imagery is one of the most efficient ways to acquire large-scale or even global observations of the Earth's surface.
Synthetic Aperture Radar (SAR), as one typical sensor for Earth Observation, emits electromagnetic waves and measures amplitude and phase of the signal scattered on the ground (using different polarisations in the case of polarimetric SAR (PolSAR)).
As an active sensor, it is independent of daylight. Due to the fact that the used microwaves are capable of penetrating fog, clouds, etc, it is less influenced by complex weather conditions and environmental changes.
Its applications include urban planning~\cite{duan2017improved}, forest protection~\cite{antropov2011volume}, the monitoring of agriculture~\cite{hajnsek2009potential}, oceans~\cite{huang2009fast,wang2016sea}, and natural hazards~\cite{tralli2005satellite}, as well as land use and land cover (LULC) classification which aims to assign a label from a set of predefined semantic categories to every pixel in the image~\cite{Shermeyer2020}.
\IEEEpubidadjcol

Methods for semantically segmenting PolSAR images can be roughly categorized into three groups: i) Approaches modelling the statistical characteristics of different backscattering types dominating the different LULC classes, ii) approaches that apply machine learning-based classifiers after extracting hand-crafted features, and iii) approaches that aim to compute such features automatically from the given data (e.g. deep learning). 
One of the most well-known statistical modelling approaches is the Wishart classifier~\cite{lee1994classification}, i.e. a maximum likelihood classifier based on the assumption that the local polarimetric covariance matrices in a homogeneous area follow a complex Wishart distribution. 
While being based on a clear mathematical framework, the model requires simplifying assumptions to estimate its parameters (e.g. fully developed speckle and locally homogeneous areas). In particular with increasing resolution, these assumptions tend to fail and introduce errors. More sophisticated approaches (see e.g. \cite{Deng2017} for a good overview and a discussion) are able to model more complex distributions. This, however, comes at the cost of increasing difficulty to accurately estimate the models parameters.

The second type extracts various (polarimetric) features, e.g. by applying target decomposition theorems~\cite{cloude1996review,yamaguchi2005four} or by computing color and texture descriptors~\cite{cheng2015segmentation}, which are then fed into typical machine learning classifiers such as Support Vector Machines~\cite{fukuda2001support,He2013} or Random Forests~\cite{Du2015}. 
This discriminative approach focuses on modelling the decision boundary or posterior distribution of different classes instead of estimating a generative model. 
Modern approaches combine traditional features with sophisticated feature selection techniques (e.g. casting the selection as optimization problem with sparsity regularization \cite{Huang2021}), directly rely on more data-driven methods for feature extraction (e.g. using sparse dictionaries of local image content \cite{Wang2021}), or combine feature extraction with deep learning (e.g. using optimized data representations and attention modules as in \cite{Jing2021}). 
While having been the virtual standard in the field for years, these approaches suffer from a strong dependency on the usefulness of the extracted image features. The successful selection of features that are meaningful for a given combination of sensor, data, and semantic categories requires strong expert knowledge in several different scientific domains.

The third type of approach aims to mitigate this limitation by working on the PolSAR data directly (or a low-level representation such as polarimetric covariance matrices) and automatically derives features that are optimal for a given task. 
The most well known example of such approaches are deep neural networks using either a real-valued vector representation of the polarimetric covariance matrix as input~\cite{zhou2016polarimetric} or directly the complex-valued data by employing a complex-valued network architecture~\cite{Zhang2017}. 
Deep learning methods have been adopted to various SAR data types and applications and are used with growing success (see e.g.~\cite{Zhu2021} for an excellent overview). 
However, they commonly require large-scale datasets for training to prevent overfitting and to reach a satisfactory performance. 
In Remote Sensing in general and for PolSAR data in particular, such large scale datasets are scarce as manual annotation is tedious and usually has to be performed by experts. 
One possibility to address the problem of data scarcity is pre-training or self-supervised learning, i.e. learning a significant part of the network parameters on a proxy task. One the one hand, this proxy task needs to be sufficiently similar to the target task rendering the extracted features meaningful. On the other hand, it needs to provide more available data. 
An example is SAR-to-optical transcoding, i.e. producing an optical look-a-like from a given PolSAR image and using the extracted features as input to a classification network~\cite{ley2018gansar}.

Another possibility is to use a special kind of shallow learners such as Random Forests (RFs) (see e.g.~\cite{Criminisi2012} for an extensive overview) that while still being able to perform automatic feature learning do not have such strong requirements on the availability of large-scale training sets. 
Random Forests are one of the most powerful shallow learners~\cite{Caruana2006} with a large body of theoretical work analyzing their properties (e.g.~\cite{Biau2012}), practical considerations such as efficiency~\cite{sharp2008implementing,haensch2018fastertrees} and interpretability~\cite{Haensch2019}, as well as applications such as feature ranking~\cite{Verikas2011}, image matching~\cite{Lepetit2006} and classification~\cite{Bosch2007}, as well as semantic segmentation~\cite{Shotton2008}.
They were extended in~\cite{hansch2018skipping} to work directly on PolSAR images and similar to Deep Learning automatically learn meaningful features. 

Despite the success and widespread application of RFs, it has been questioned whether their hierarchical arrangement of binary features is responsible for their good performance~\cite{ozuysal2007fast}. 
As an alternative, Random Ferns (RFe) have been suggested. Similar to RFs, they compute simple binary features but forgo the feature hierarchy and instead follow a semi-naive Bayesian approach to arrange features in groups. 
Such a feature group, called fern, replaces the single trees of RFs and acts as weak base-learner which estimates the full joint class posterior given its features. 
In contrast to RFs, the individual estimates of the weak learners are not combined by an (additive) average, but via multiplication. 
This flat structure allows a much simpler model in terms of memory requirements and computational costs without degrading the discriminative power. 

The term "Random Ferns" (RFe) has been coined in~\cite{ozuysal2007fast,Ozuysal2009} which proposed them as a computational more efficient alternative to an earlier approach relying on Random Forests to perform keypoint matching for photogrammetric computer vision tasks such as pose estimation of planar surfaces and panorama stitching~\cite{Lepetit2006}. 
The computational efficiency of RFe was further shown in~\cite{Wagner2008} for pose tracking on mobile phones. 
The work shows that RFe can outperform more complex approaches while being based on much simpler operations. 
Other works include e.g. image-based localization~\cite{Donoser2014}, pose estimation~\cite{Krupka2014,Hesse2015}, tracking~\cite{Godec2013,Kwak2017}, and general object detection~\cite{Villamizar2010} potentially combined with online learning \cite{Villamizar2012c}.
In~\cite{Bosch2007} RFe are used for image classification, i.e. assigning a class label to the image as a whole, and compared against Random Forests and a multi-way SVM. 
Experiments show that all three approaches led to similar results with RFe being computationally much more efficient than either, RF or SVM, and - together with a novel pyramidal feature representation - outperform the state of the art of the time by a large margin. 
In~\cite{Kursa2014} RFe are extended to a general-purpose machine learning algorithm. 
Furthermore, several concepts originally defined for RFs are adapted for RFe including an internal error approximation as well as an attribute importance measure.
The first description of using RFe for semantic segmentation, i.e. the pixel-wise semantic annotation of images, can be found in~\cite{Gonfaus2012TowardsDI} for natural images. 
A similar approach albeit in the context of medical image processing is used for the segmentation of kidney components in CT data~\cite{Jin2017} and cells in microscopic images~\cite{Browet2016}.
In contrast to Random Forests (see e.g. \cite{Belgiu2016} for an overview), Random Ferns have so far been rarely used in remote sensing applications. 
To the best of the authors' knowledge, the proposed approach is the first that applies RFe in the context of semantic segmentation of remote sensing imagery in general and for PolSAR data in particular. 

Different from Random Forests, the original formulation of Random Ferns does not include a built-in feature selection. 
Instead, similar to Extremely Randomized Trees~\cite{Geurts2006}, the computation of the internal binary features relies on parameters that are sampled completely randomly. 
One of the few approaches to optimize RFe is proposed in~\cite{Villamizar2010} and relies on boosting, i.e. the adaptive change of sample weights based on initial estimates of the classifier. 
Originally limited to binary problems, it was later extended in~\cite{Sharma2014} to multi-class tasks and applied for object detection in images~\cite{Villamizar2012a,Villamizar2017boosted}. 
Another line of work uses stacking~\cite{Wolpert1992}, similar to stacked Random Forests~\cite{Zhou2017,Haensch2018}, and groups multiple RFe classifiers in layers where each layer obtains the original image data as well as the estimate of the previous layer as input~\cite{Kim2019,Kim2020}. 
The approach has been evaluated for handwritten digit recognition and face recognition and shows competitive performance to the state of the art based on deep learning while having significantly less parameters to optimize. 

In this paper, we make three novel main contributions: 
First, all of the previous RFe approaches are designed for real-valued data, either feature vectors ($\mathbb{R}^n$) or optical images ($\mathbb{R}^{N_x \times N_y \times N_s}$, where $N_x \times N_y$ is the spatial image dimension and $N_s$ the number of spectral channels). 
PolSAR images, however, are complex valued data containing either the polarimetric scattering vector or the Hermitian polarimetric covariance matrices. 
In this paper we extend the original RFe framework by adapting the RF node projections of~\cite{hansch2018skipping} as the ferns' binary features. 
This allows to apply the proposed RFe directly to PolSAR images without the need of explicitly computing hand-crafted features. 
Second, we address the lack of optimization in the original RFe formulation and propose two novel optimization strategies to further increase their performance. 
They do not require an iterative update of sample weights (as boosting does) making them less prone to label noise, lead to a single optimized RFe classifier (in contrast to stacking) minimizing the computational load, and maintain independence between different classifier parts (in contrast to both, stacking and boosting) allowing for efficient parallel computations. 
While they do require a longer training time, the inference time remains unchanged which is more critical in most applications.
While the first contribution allows omitting the extraction of hand-crafted features and applying the proposed framework to the complex-valued PolSAR data directly, the second contribution enables the RFe to automatically learn features from the data in terms of an optimized combination of optimized binary features. 

Third, we apply the proposed framework to the semantic segmentation of PolSAR imagery and evaluate its performance on a modern dataset. 
The high efficiency of RFe enables their use for semantic segmentation of large scale image data. 
The experiments show that the proposed classifier achieves similar accuracy to RFs despite being a much simpler model and is on par with a much more sophisticated deep learning model.

In summary, we make the following contributions to the state of the art:
\begin{itemize}
    \item We extend RFe to be applied to PolSAR images directly, i.e. to complex-valued image data where each pixel contains a Hermitian matrix. This mitigates the requirement of other shallow learners to be trained on predefined real-valued image features. Instead, it enables automatic feature learning similar to corresponding RFs approaches~\cite{hansch2018skipping} or deep learning.  
    \item We propose two novel approaches to boost the classification performance of RFe by optimizing the groups of internal binary features. While this comes at a higher computational cost during training, the resulting classifier remains efficient during inference.
\end{itemize}

The remainder of this paper is organized as follows: Section~\ref{sec:section2} gives a brief introduction of PolSAR data and polarimetric distances since they are needed for computing the binary features. Section~\ref{sec:section3} briefly explains the Random Fern classifier and introduces the proposed binary features. 
Section~\ref{sec:section4} presents the experiments and discusses the obtained results. 
Section~\ref{sec:section5} concludes the paper and gives an outlook on future work.

\section{Polarimetric Synthetic Aperture Radar\label{sec:section2}}

Polarimetric SAR systems emit polarized microwaves and measure the amplitude and phase of the part that is scattered on the ground back to the sensor. 
A PolSAR measurement can be represented as a $2\times2$ complex-valued scattering matrix $\mathbf{S}$ that contains the polarimetric scattering characteristic of the target object on the ground, i.e.
\begin{equation}
\mathbf{S}=\left(\begin{array}{cc}
S_{H H} & S_{H V} \\
S_{V H} & S_{V V}
\end{array}\right),
\label{eqn:scatteringmatrix}
\end{equation}
where $S_{TR}\in\mathbb{C}$ denotes the complex scattering information, $T,R\in\{H,V\}$ is the polarization of the transmitted and received signal, respectively, and $H$ and $V$ denote horizontal and vertical wave polarization states.
Under monostatic backscattering the cross-polarization terms have the same scattering information, i.e. $S_{HV} = S_{VH}$, and the scattering matrix $\mathbf{S}$ can be expressed as the lexicographic scattering vector
\begin{equation}
\mathbf{k_L}=\left(S_{H H}, \sqrt{2} S_{H V}, S_{V V}\right)^{T}.
\end{equation}

As distributed targets can only be fully described by second-order moments, an often used data representation is the local variance-covariance matrix $\mathbf{C}$ obtained by locally averaging the outer product of the scattering vectors, i.e. 
\begin{equation}
\mathbf{C} =\left\langle\mathbf{k_L}\mathbf{k_L^\dag}\right\rangle,
\label{eqn:covariance}
\end{equation}
where $\left\langle\cdot\right\rangle$ and $\cdot^\dag$ denote spatial averaging and complex-conjugate transpose, respectively.

While there are many approaches to extract information from a polarimetric covariance matrix $\mathbf{C}$ (e.g. polarimetric decompositions~\cite{Alberga2004}, scalar features such as entropy, $\alpha$, anisotropy~\cite{Cloude2002}, or various distance measures~\cite{haensch2018poldist}), we focus on the span~$s$, i.e. the total power of the reflected signal, 
\begin{equation}
s = tr(\mathbf{C}), 
\end{equation}
where $tr(\cdot)$ is the trace operator, and the log-Euclidean distance \cite{arsigny2006log} between two Hermitian matrices $\mathbf{A}, \mathbf{B}$ defined as
\begin{equation}
d_{le}(\mathbf{A}, \mathbf{B})=\|\log (\mathbf{A})-\log (\mathbf{B})\|_{F}
\end{equation}
where $||\cdot||_F$ is the Frobenius norm and $\log(\cdot)$ denotes the matrix logarithm.
The matrix logarithm can be pre-computed which makes it possible to calculate this distance very efficiently. It has also shown excellent performance in various tasks of analyzing PolSAR imagery~\cite{Yang2018,haensch2018poldist}.

\section{Methodology\label{sec:section3}}
\subsection{Random Ferns Classifier}
A Random Ferns (RFe) classifier as proposed in~\cite{ozuysal2007fast} is an ensemble approach. 
Similar to Random Forests (RFs, see e.g.~\cite{hansch2018skipping}), a set of weak classifiers is trained by evaluating binary features sampled from a large (i.e. potentially infinite) feature pool. 
The final classification decision is determined by aggregating the results of the individual weak learners. 
The two largest differences to RFs are that 1) features are not hierarchically ordered in a tree structure, i.e. every feature is computed for every sample, and 2) the weak learners are combined via multiplication rather than averaging.

In this section, we briefly repeat the definition of RFe to introduce the used notation, then focus on the particularities regarding analyzing PolSAR images in Section~\ref{sec:polsar-feat}, and close with the introduction of two novel approaches to optimize RFe in Section~\ref{sec:fernopt}. 

Given a set $D$ of training samples ${({\mathbf p},c) \in D \subset \mathcal{P}\times \mathcal{C}}$ of input data $\mathbf p$ and class label ${c \in \mathcal{C}=\{1,2,...,L\}}$, let  ${F=\{f_i\in\{0,1\}|i=1,2,...,N\}\subset\mathcal{F}}$ be a set of $N$ (binary) features taken from all possible features $\mathcal{F}$ computed over ${\mathbf p}$. 
Classifying a sample ${\mathbf p}$ aims to find the label $c^*$ such that
\begin{equation}
c^* = \arg\max_c \left[P(c|{\mathbf p}) =  P(c|\mathcal{F}) \approx  P(c|F) \propto P(F|c)P(c)\right].
\end{equation} 

Modelling the class likelihood $P(F|c)$ as full joint probability over $N$ binary features would require to estimate $2^N$ parameters per class. This is infeasible even for moderate $N$. 
Naive Bayes~\cite{Hand2001} would assume complete statistical independence, leading to 
\begin{equation}
    P(F|c) = P(f_1, f_2, ..., f_N|c) = \prod_{i=1}^N P(f_i|c)
\end{equation}
and thus decreasing the number of parameters to $2N$ at the cost of being unable to model any dependency relationships among the features. 
RFe represent a trade-off between both extremes by assuming statistical independence between groups of features while modelling the joint probability within each group, i.e.
\begin{equation}
    \label{eq:fern}
    P(F|c) = P(f_1, f_2, ..., f_N|c) = \prod_{j=1}^M P(F_j|c)
\end{equation}
where $F_j = \{f_{j,k}\}_{k=1,...,N_j}\subset F$ (e.g. $N_j=\bar N = N/M$), $\bigcup_{j=1}^M F_j = F$, and $\forall r,s\in[1,M],r\ne s: F_r \cap F_s = \emptyset$. 

Ideally, features located within the same group share dependencies, while the different groups are statistically independent. 
In reality, this assumption will most likely be violated in the same way as the statistical independence assumption of Naive Bayes. 
However, it allows tractable training and inference (i.e. it has $\sum_{j=1}^M 2^{N_j} \approx M\cdot 2^{N/M}$ parameters). On the other hand, it is still able to model certain feature interactions since each feature group $F_j$ models a full joint probability over $N_j$ binary features. 
Note, that a RFe reduces to the Naive Bayes classifier (based on binary features) for $M=N$, while for $M=1$ the full joint probability is modelled - thus, RFe scale between these two extremes.

In practice, the joint probability $P(F_j|c)$ is represented by an $N_j$-dimensional histogram per class. 
Since every dimension (being based on binary features) has only two possible values, the bins can easily be enumerated and indexed by ${l_j=\sum_{k=1}^{N_j} 2^{k-1}f_{j,k}}$. 
This allows implementing RFe as highly efficient look-up tables. 
During training, all features $F$ are computed for every sample $\mathbf p$ and the $l_j$-th bin in each fern is increased according to the class given by $c$, i.e. the label of training sample $\mathbf p$.

During prediction, the same $N$ features within the $M$ feature groups are evaluated for a given query sample $\mathbf{p}$ providing the leaf index $l_j$ for each fern $F_j$. 
Laplace smoothing is applied while normalizing the learned absolute class histograms $\hat P(F_j|c)$ to represent class conditional probabilities $P(F_j|c)$, 
\begin{equation}
    P(F_j|c) = \frac{\hat P(F_j|c)+u}{\sum_{l_j}(\hat P(F_j|c)+u)}
\end{equation}
where $u$ is a constant. The main reason for this correction is that if a feature combination within a single fern never occurred during training, the corresponding class likelihood is estimated as $P(F_j|c) = 0$. If it is queried during prediction, it will cause $P(F|c)=0$ independent of the other ferns' estimates since the individual predictions are aggregated by multiplication (see Eq.~\ref{eq:fern}).

\subsection{Binary Features for PolSAR Data}
\label{sec:polsar-feat}
RFe have originally been proposed for color images, (i.e. arrays of real-valued 3D vectors)~\cite{ozuysal2007fast}. 
In this case, the binary features are comparisons of pixel values within a given patch where pixel position and color channel are randomly selected. 
Inspired by~\cite{hansch2018skipping}, we extend this concept to polarimetric SAR images by using patch projections directly defined over the space of arrays of Hermitian matrices. 

We define a binary feature~$f$ as a distance~$d$ between Hermitian matrices which is then compared to a scalar threshold~$\delta$, i.e. 
\begin{equation}
f=\left\{\begin{array}{l}
1 \text {, if } \hat p\geq\delta \\
0 \text {, otherwise,}
\end{array}\right.
\end{equation}
where $\hat p = d\left(\psi(\phi_1(\mathbf{p})),\psi(\phi_2(\mathbf{p}))\right)$.
The projection that computes a real-valued scalar $\hat p$ from a patch $\mathbf{p}$ of Hermitian matrices applies an operator $\psi(\cdot)$ to two patch regions that are randomly selected by $\phi(\cdot)$ around the center of the query patch~$\mathbf{p}$. 
There are several choices for $\psi$, e.g. computing the average value of the region, using the value at the region center, or selecting the value with minimal or maximal span within the region \cite{hansch2018skipping}. 

\begin{figure}
\begin{minipage}[t]{0.24\textwidth}
\centering
\includegraphics[width=\textwidth]{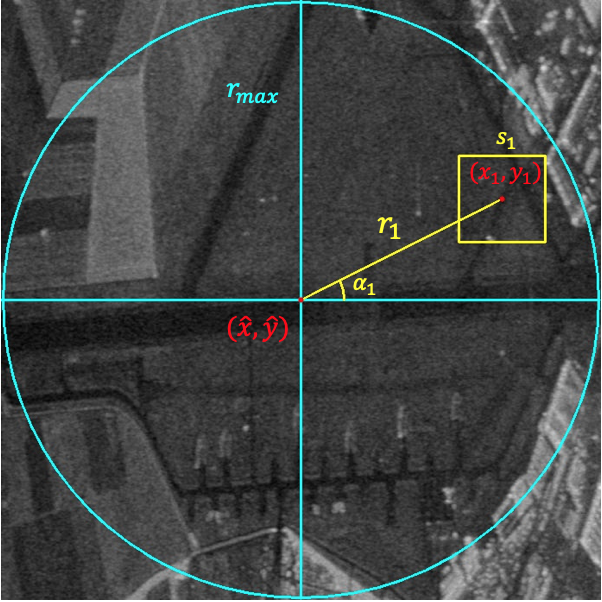}
\subcaption{\centering One-point projection}
\end{minipage}
\begin{minipage}[t]{0.24\textwidth}
\centering
\includegraphics[width=\textwidth]{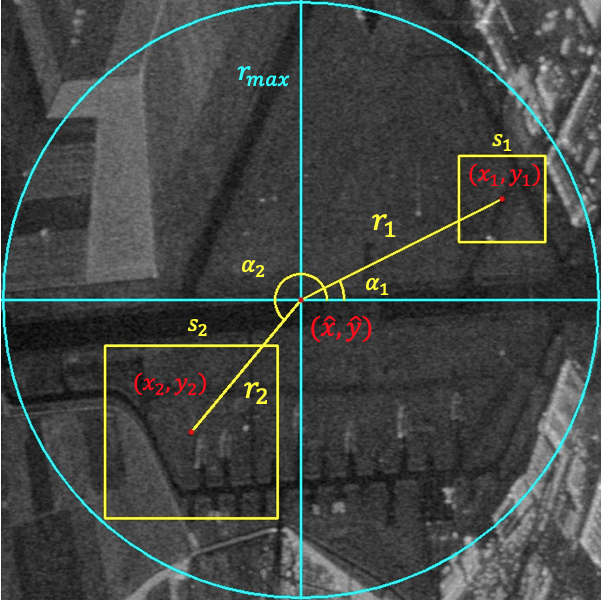}
\subcaption*{ \centering Two-point projection}
\end{minipage}
  \caption{We use two types of projections: One-point projections (left) compare a region randomly selected within the patch with a reference pixel randomly chosen from the training samples. Two-point projections compare two regions both randomly selected within a patch. A region is defined by its size~$s$ and position~$(x,y)$ which is computed with a random offset vector from the patch's $\mathbf{p}$ center~$(\hat{x},\hat{y})$ sampled in polar coordinates, i.e. having a distance $r<r_{max}$ and orientation~$\alpha$.}
  \label{fig:projection}
\end{figure}

We consider two different types of such projections (illustrated in Figure~\ref{fig:projection}) depending on whether the second region $\phi_2(\mathbf{p})$ is sampled within the patch or from the whole image. 
The latter is a simple way to create a random value that follows the data distribution of the given image and thus analyzes absolute patch properties (e.g. intensity, backscatter type, etc.). 
The former analyzes relative properties of pixels within the patch such as local texture.

The corresponding projection function depends on several parameters which are randomly selected from a uniform distribution $U$ with user-defined ranges: Whether a one- or two-point projection is used, the region size~${s\sim U\left(1, s_{max}\right)}$ and orientation angle~${\alpha\sim U(0,360)}$ as well as distance ${r\sim U\left(0, r_{max}\right)}$ to the patch center, and the threshold ${\delta\sim U(\min_D(\hat p), \max_D(\hat p))}$. 
While there are different potential choices for the region operator~$\psi$ as well as the polarimetric distance function~$d$, we settle for selecting the region pixel with maximal span and the log-Euclidean distance (see Section~\ref{sec:section2}). 

\subsection{Fern Optimization}
\label{sec:fernopt}
In the original RFe formulation~\cite{ozuysal2007fast}, full randomization is used during the selection of the binary features as well as during their grouping. 
This is in contrast to 
RFs which create random subsets of possible splits within each internal node and select the best based on certain optimization criteria (e.g. using the information gain in Eq.~\ref{eq:infogain}). 
We propose two different optimization methods for RFe, which can be applied independently or jointly. While the first approach (Section~\ref{sec:presel}) is a pre-processing step before the actual creation of the RFe, the second method (Section~\ref{sec:itopt}) is an iterative procedure based on an already existing RFe classifier.

\subsubsection{Preselection and grouping of binary features}
\label{sec:presel}
There are two main issues with the standard approach of full randomization during feature selection and feature grouping. 
First, the overall set of potential features includes many that are meaningless - either in general (e.g. the difference between values at identical pixel positions) or for the specific classification task. 
Selecting these features leads to uniform distributions in the corresponding parts of the class posterior and thus only adds computational load without an increase in performance. On the other hand, the effective number of informative features decreases. 
One could attempt to avoid the creation of meaningless features by hard-coding constraints during feature selection (e.g. ensuring not to select identical pixel positions). However, these rules quickly become very complicated and numerous to catch all possible exceptions, add a computational burden, and are not effective to detect features that are not meaningless in general, but not informative for the given task.
A more suitable approach is to create a large pool of feature candidates and select the ones with a high potential to be descriptive.
The second issue is that the potential set of features contains many that are redundant - again either in general (e.g. features with identical values) or for the task at hand. 
While one could attempt to define manual rules to avoid the selection of highly similar features, this approach is futile for the same reasons as stated above.

Instead, we propose an automatic feature selection approach similar to the node optimization in RFs and extend it with a grouping approach. 

A single binary feature divides the feature space into two half-spaces. 
Each of them holds samples that are - with respect to this feature - more similar to each other than to samples in the other group. 
In the optimal case, samples of a given class are only in one part but not in the other which would lead to pure class posteriors. 
We use measures of impurity to judge the quality of the corresponding binary feature, i.e. the more a binary feature decreases the impurity, the more discriminative is it.

Inspired by the node optimization in RFs, we use an adapted version of the information gain~$IG$ which measures the difference between the class impurity~$I$ before and after a split, i.e.
\begin{equation}
\label{eq:infogain}
    IG(f) = I(D)-P_0\cdot I(D_0)-P_1\cdot I(D_1)
\end{equation}
where $I$ is a measure of impurity, $D$ is the given dataset, ${D_k=\{\mathbf{p}\in D|f(\mathbf{p})=k\}}$ and ${P_k = |D_k|/|D|}$ (with ${k\in\{0,1\}}$).

We use the entropy of the class posterior as an impurity measure~$I$. 
While $D$ is the local subset of the training data at an internal node in the case of RFs, it corresponds here to the total set of training samples. Its impurity is therefore constant for all binary features and can be ignored. 
Thus, instead of selecting features that maximize Eq.~\ref{eq:infogain}, it is sufficient to select features that minimize
\begin{equation}
    \hat{IG}(f) = P_0\cdot I(D_0)+P_1\cdot I(D_1).
\end{equation}

It should be stressed that this is a greedy optimization procedure which only takes single features into account. 
A feature is rejected if it leads to a very impure split by itself. 
Feature combinations are not considered at all and there is no guarantee that selecting individually optimal features leads to an optimal feature set. 
That being said, while there is a risk to disregard features that are meaningful only in combination with other features, truly meaningless and very weak features are rejected as well while very strong features are selected. 

However, redundant and strongly correlated features cannot be detected by this approach. 
On the one hand, redundant features should be rejected. 
On the other hand, RFe assume mutual independence between the individual feature groups which requires placing correlated features into the same group. 

To this aim, we compute the correlation matrix between all selected binary features as
\begin{equation}
    \label{eq:corr}
    corr(f_r,f_s) = \frac{cov(f_r,f_s)}{\sigma_{f_r}\sigma_{f_s}}
\end{equation}
where $cov$ denotes the covariance between two features and $\sigma_{f}$ the standard deviation of feature $f$.
This allows us to reject highly correlated (and thus redundant) features and group correlated features together to mitigate statistical dependencies between the individual ferns.

\subsubsection{Iterative fern optimization}
\label{sec:itopt}
The approach in Section~\ref{sec:presel} selects strong binary features while rejecting redundant features and groups correlated features within one fern. 
However, it does not take the performance of feature combinations into account and does not optimize the RFe as a whole. 

In this section we propose a second, complementary optimization scheme which is inspired by Monte-Carlo type algorithms such as simulated annealing.

Starting with an initial RFe $h_0$ (created randomly or via the optimization approach presented in the previous section), we define a set of operators on the current instance $h_t$ which create a slightly changed instance $\hat{h}_t$:
\begin{itemize}
    \item Add a new feature group.
    \item Add a new feature to a feature group.    
    \item Delete a feature from a feature group.
    \item Switch two features in different feature groups.
    \item Sample a new threshold for a feature.
\end{itemize}
The feature group and the feature within a group that are affected by such a change are randomly selected. Note, that most of these changes come with only a small additional cost. In particular, it is not necessary to retrain the complete RFe but in the worst case only small specific parts of it which can be achieved quickly given the overall efficiency of RFe.

After such a random change transforms $h_t$ into $\hat{h}_t$, the latter is evaluated on a validation set. 
If the performance increased, the change is kept (i.e. $h_{t+1}=\hat{h}_t$), otherwise it is discarded (i.e. $h_{t+1}={h}_t$).
This procedure is repeated until convergence, i.e. until all changes within a predefined number of iterations have been rejected.

\section{Experiments\label{sec:section4}}
\subsection{Data \& Performance Metrics}

\begin{figure}[ht]
\begin{minipage}[t]{0.24\textwidth}
\centering
\includegraphics[width=\textwidth]{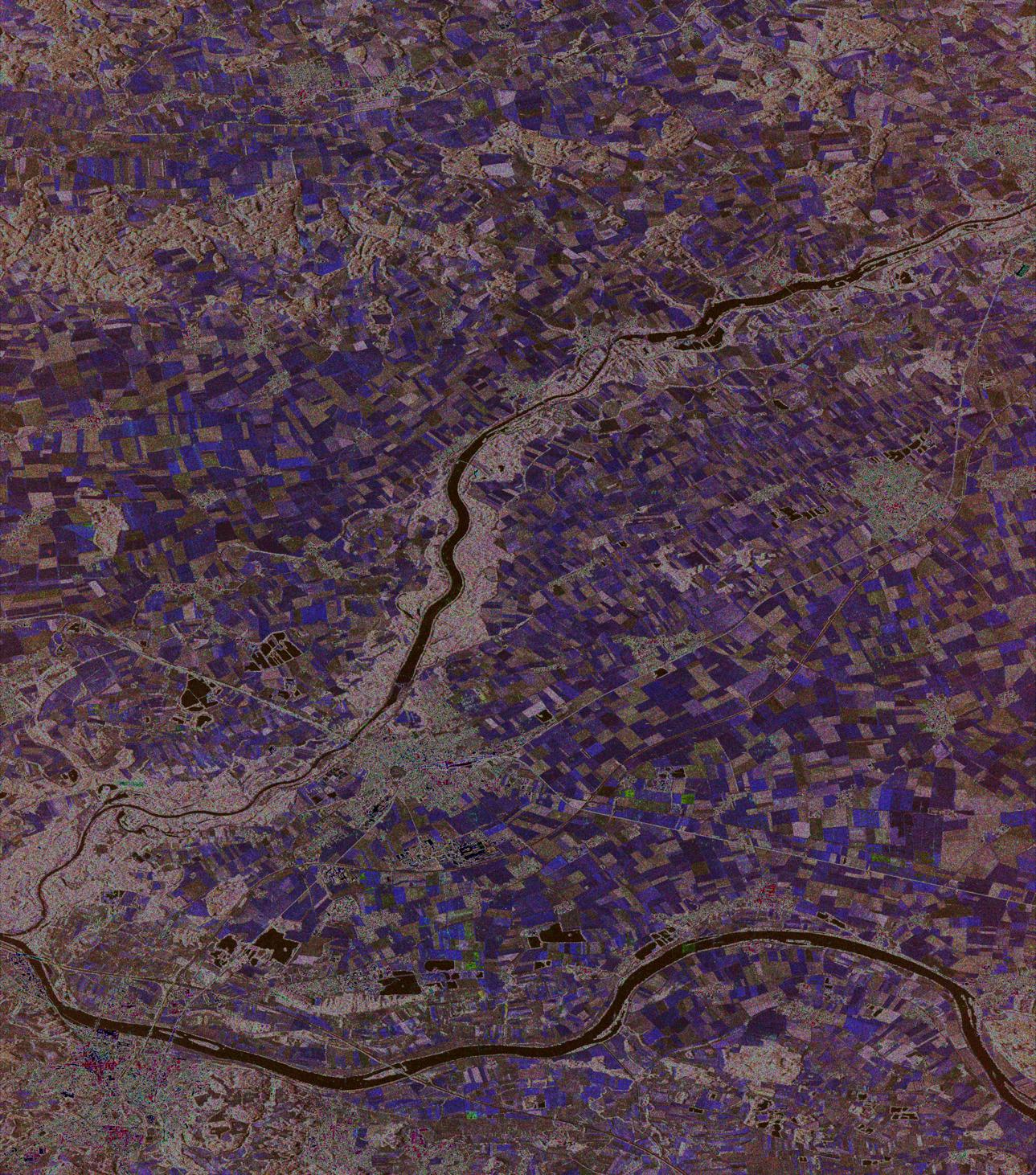}
\captionsetup{justification=centering}
\subcaption{
}
\label{fig:plattling}
\end{minipage}
\begin{minipage}[t]{0.24\textwidth}
\centering
\includegraphics[width=\textwidth]{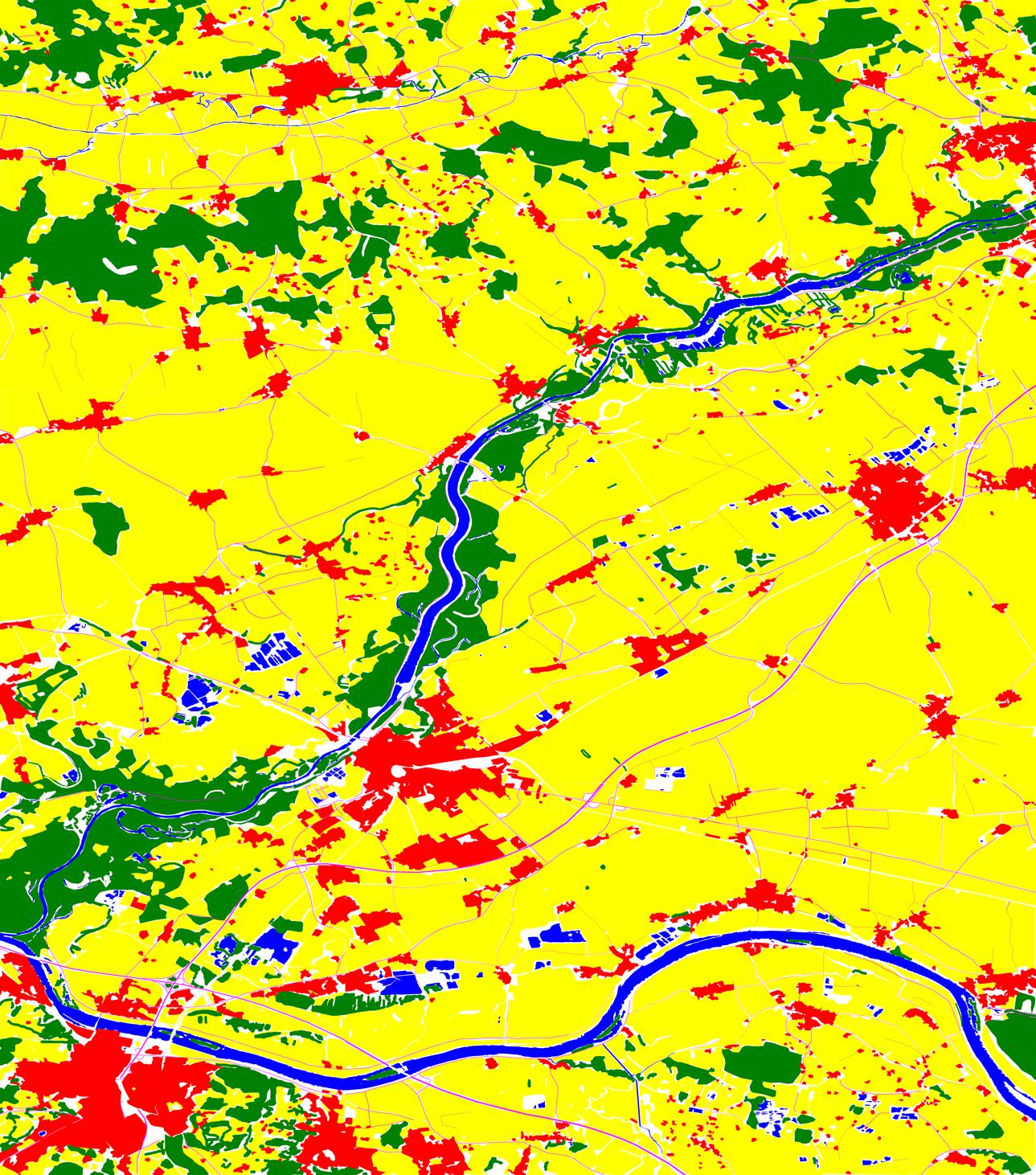}
\captionsetup{justification=centering}
\subcaption{
}
\label{fig:reference}
\end{minipage}
  \caption{(a) Pseudo-color image of TerraSAR-X data acquired over Plattling, Germany.  
  (b) Manually annotated reference data (Red: urban area, yellow: field, green: forest, blue: water, magenta: roads, white: unlabelled pixel).}
\end{figure}

\begin{figure}[ht]
\centering
\includegraphics[width=0.7\columnwidth]{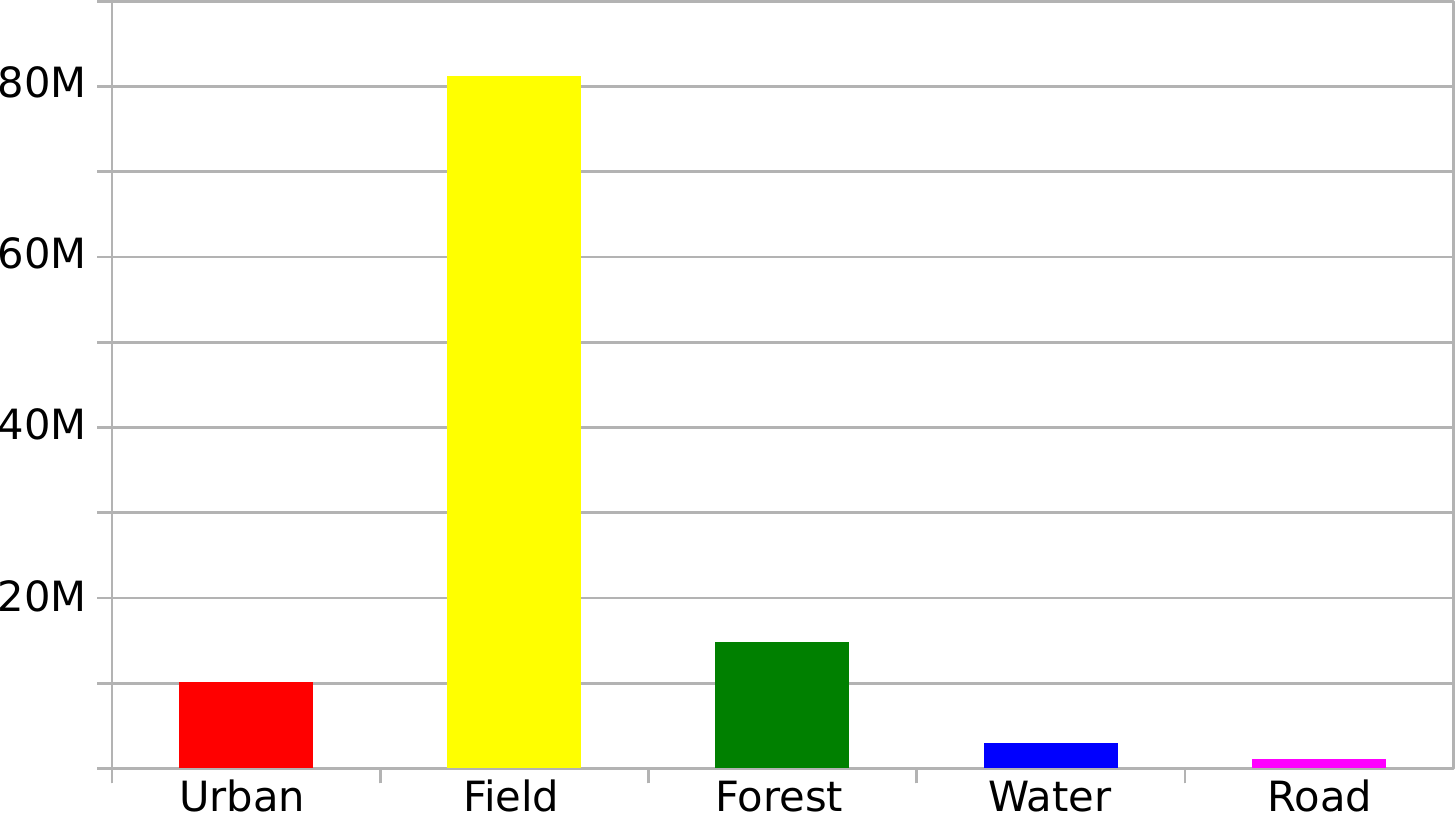}
  \caption{Distribution of manually annotated samples.} \label{fig:plattling_class_distribution}
\end{figure}

The following experiments are conducted on a fully polarimetric SAR dataset acquired by TerraSAR-X (DLR) over Plattling, Germany (a pseudo-color image is shown in Figure~\ref{fig:plattling}). 
The image contains $10312\times 11698$ pixels and was manually annotated with five predefined classes including natural media (forest, water, field) and man-made objects (urban, roads) resulting in more than 110M labelled samples. 
Figure~\ref{fig:reference} shows the obtained semantic map and Figure~\ref{fig:plattling_class_distribution} the corresponding label distribution. 
Results are based on 5-fold cross-validation, i.e. the image is divided into five disjoint stripes of equal size where training and validation sets are sampled from four stripes while the fifth stripe is used for testing. 
Additionally, the experiments have been repeated four times per fold to account for the randomness in the method (i.e. due to sampling training data and fern creation/initialization).

We employ overall accuracy (OA, percentage of correctly classified pixels), average accuracy (AA, accuracy averaged over all classes), mean Intersection over Union (mIoU), the F1 score, and the kappa coefficient averaged over the test data in the individual five folds (and runs) to quantitatively evaluate the classification performance.
Additionally, we evaluate the entropy 
of the estimated class posterior  
as a measure of certainty of the classifier in its decision.

\subsection{Results \& Discussion}

\subsubsection{Baseline}
\label{sec:baselineRes}
In a first set of experiments, we aim to establish a baseline for a RFe classifier without any of the proposed optimization techniques. These experiments shall also serve to illustrate the influence of fern number~$M$ (i.e. the number of feature groups) and fern size~$\bar N$ (i.e. the (average) number of features within a group) on the classification performance. We use a maximal region distance of $r_{max}=25$ and a maximal region size of $s_{max} = 9$. The RFe is trained with $3,000$ samples per class.

\begin{figure}
    \begin{minipage}[t]{0.24\textwidth}
    \centering
    \includegraphics[width=1\textwidth]{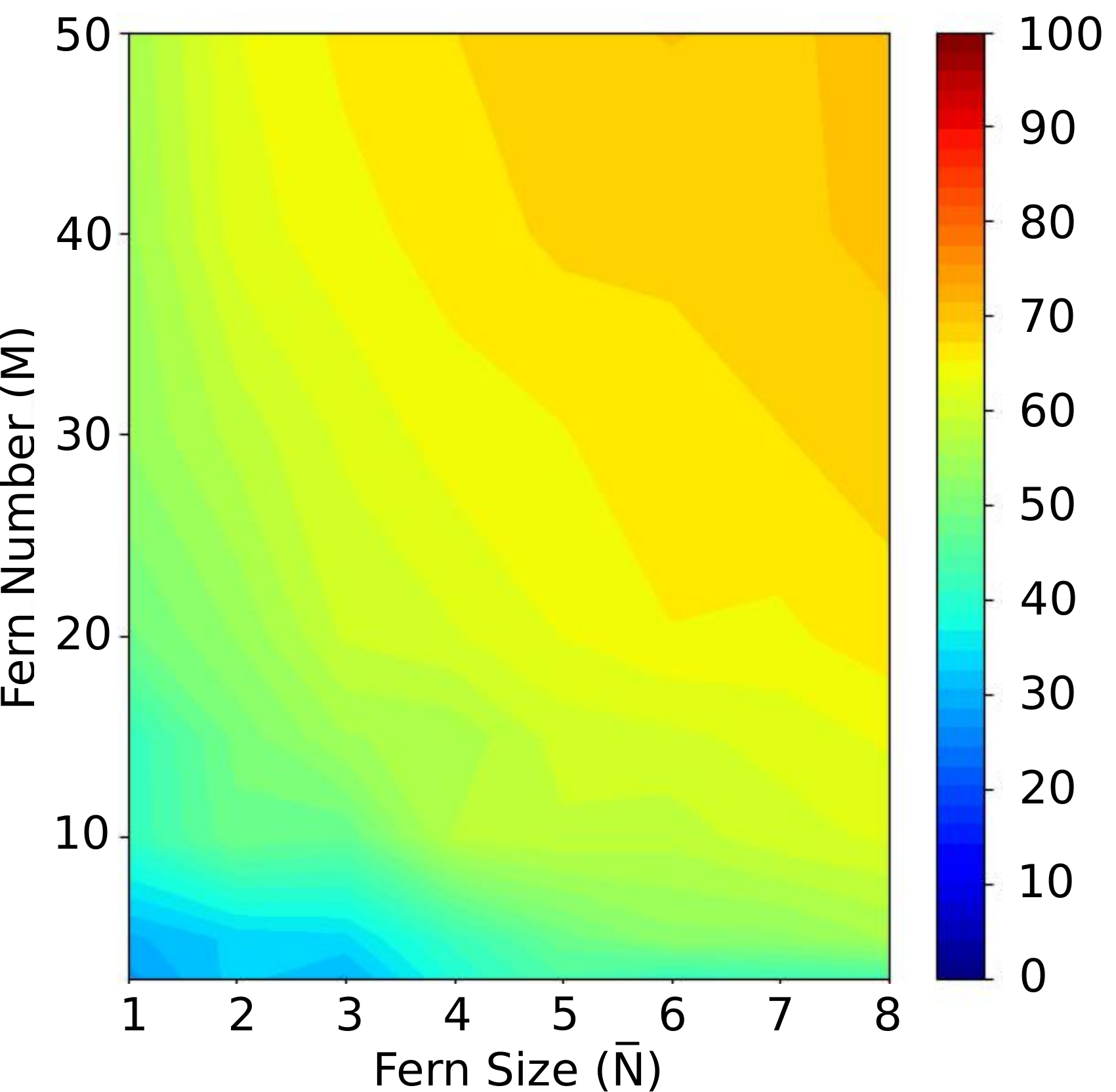}
    \subcaption{Average Accuracy\\$[0.28, 0.7]$}
    \captionsetup{justification=centering}
    \label{fig:train_contour_fernNumberSize}
    \end{minipage}
    \begin{minipage}[t]{0.24\textwidth}
    \centering
    \includegraphics[width=1\textwidth]{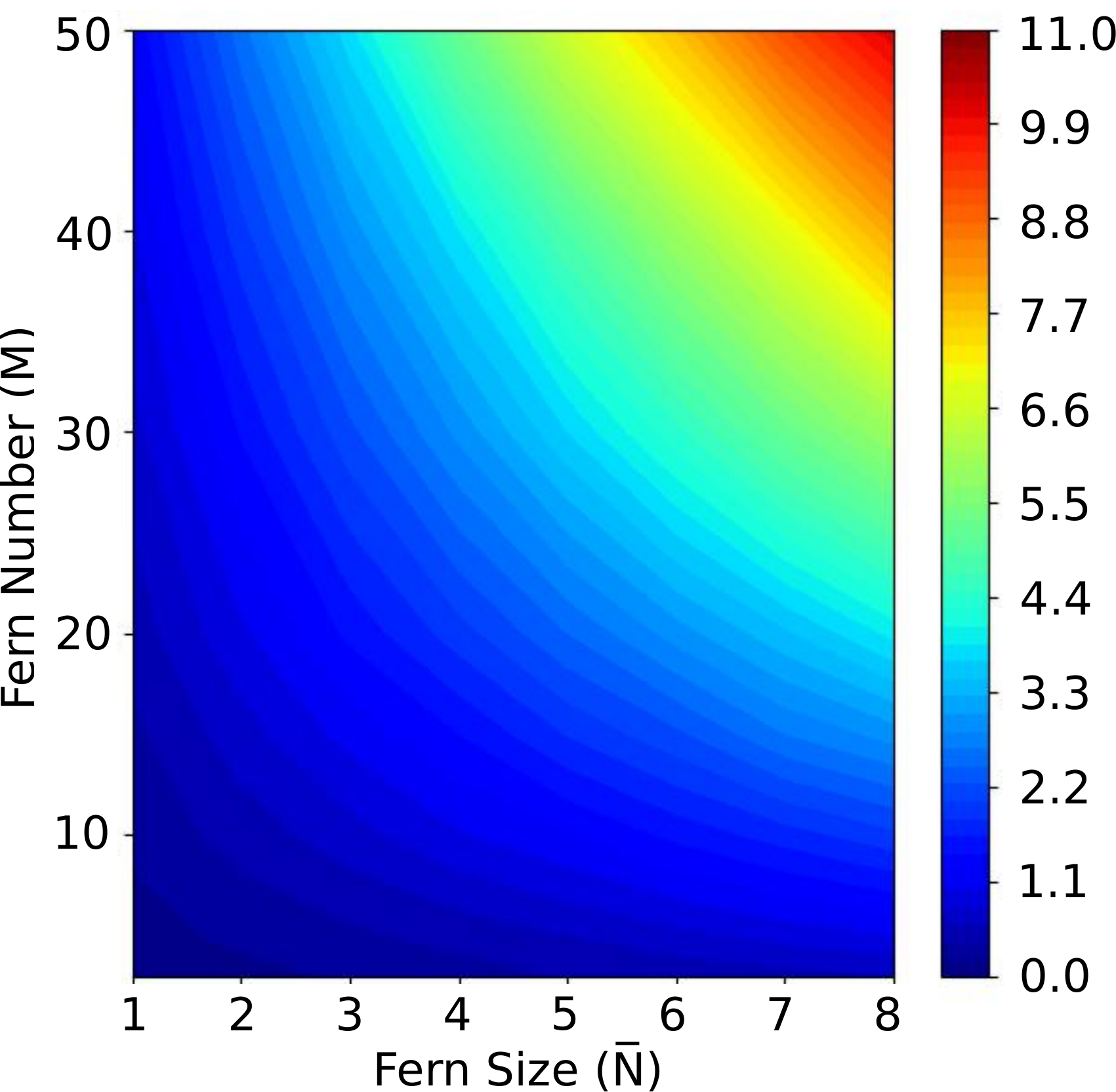}
    \captionsetup{justification=centering}
    \subcaption{Mean training time \\ $[0.08s, 10.1s]$}
    \label{fig:time_contour_fernNumberSize}
    \end{minipage}
  \caption{
  Estimated average accuracy (left) and training time (right) for RFe with different fern number and size.
  }
  \label{fig:baseresults}
\end{figure}

\begin{figure}
\begin{minipage}[t]{0.24\textwidth}
\centering
\includegraphics[width=\textwidth]{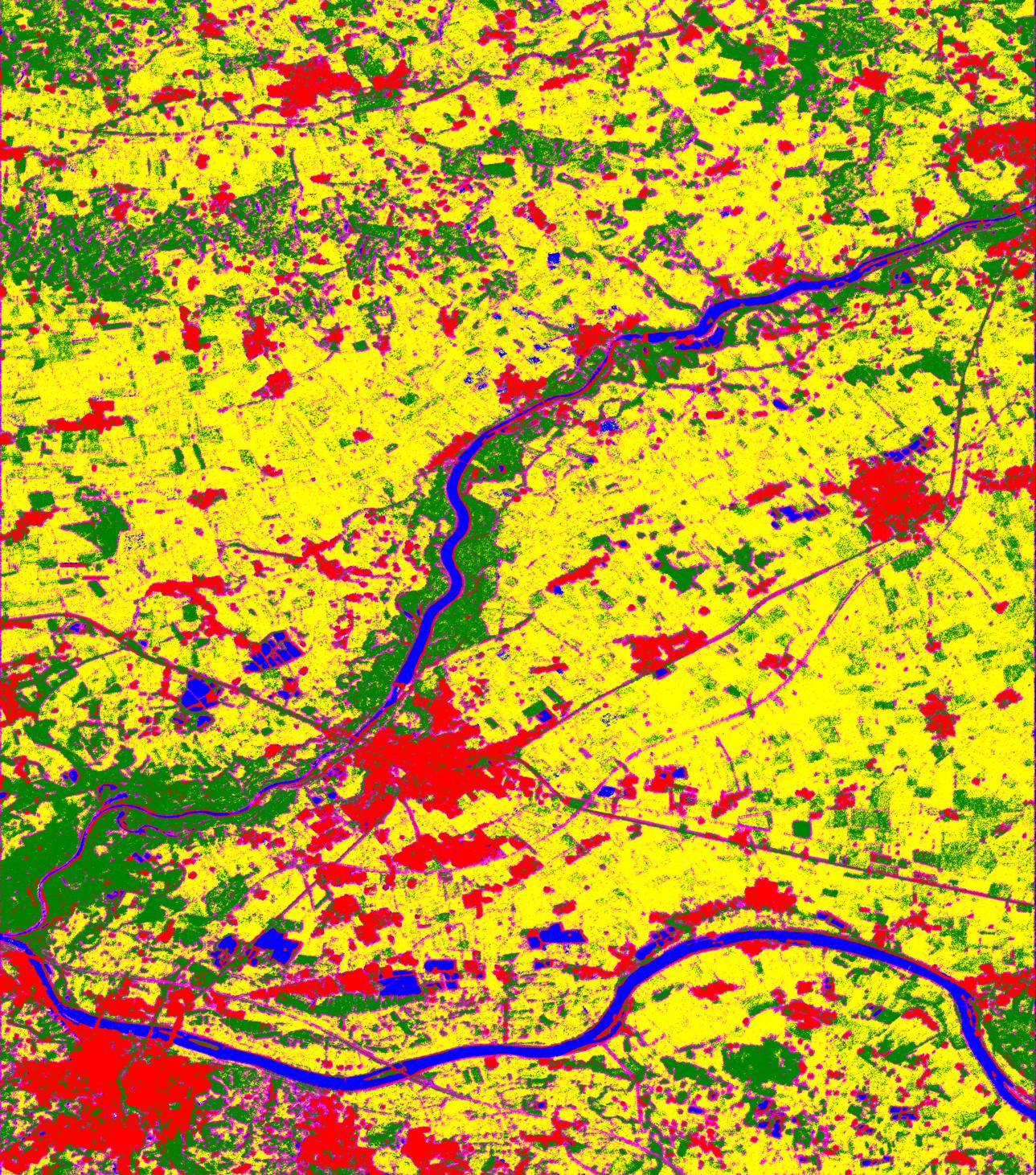}
\captionsetup{justification=centering}
\end{minipage}
\begin{minipage}[t]{0.24\textwidth}
\centering
\includegraphics[width=\textwidth]{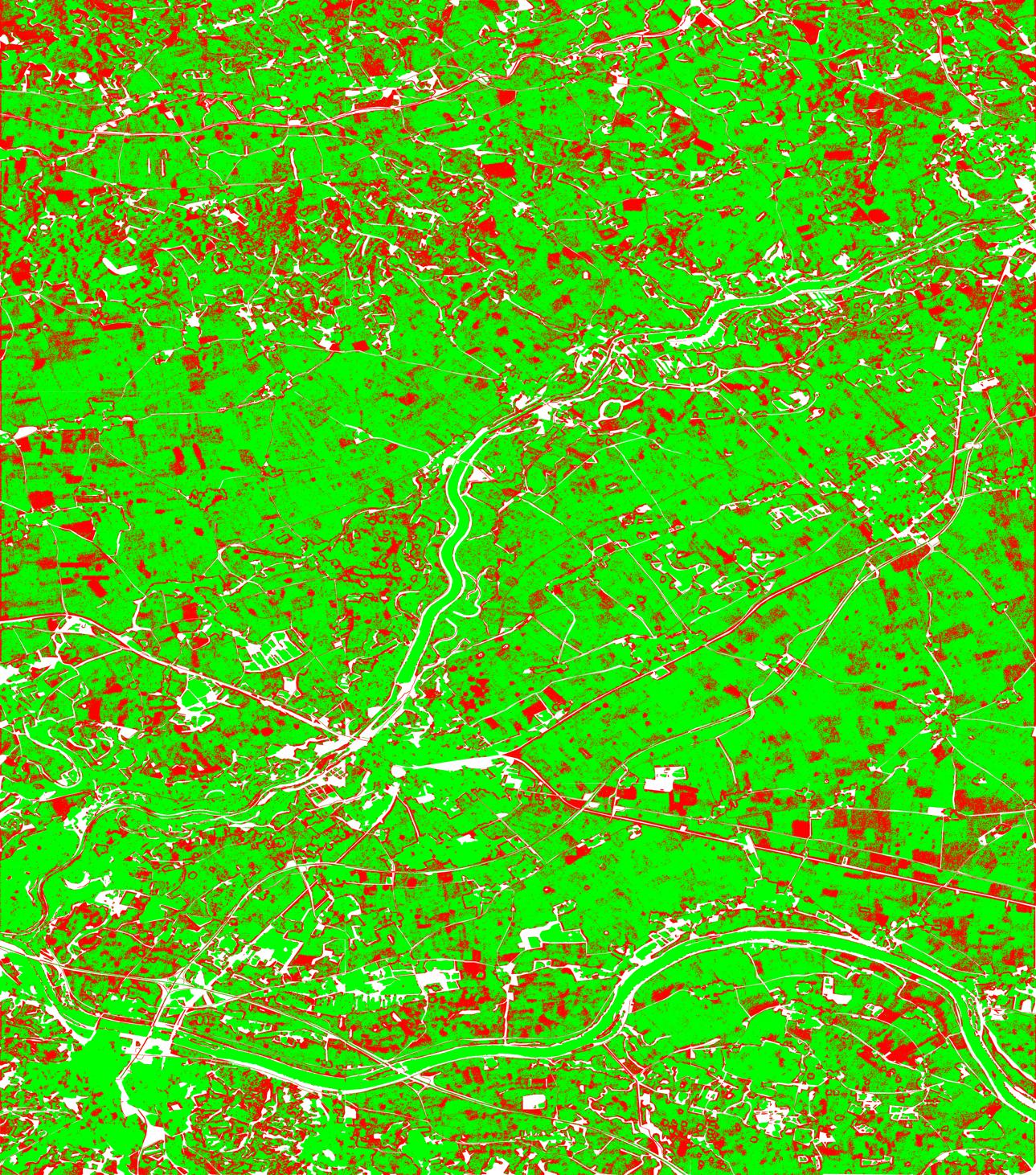}
\captionsetup{justification=centering}
\end{minipage} 
  \caption{Left: Semantic map obtained by RFe ($M=30, \bar N=8$) without optimization. Right: Error map (correct estimated label denoted in Green, wrong decisions in Red.).}
  \label{fig:base-res}
\end{figure}

The results of the conducted experiments are shown in Figure~\ref{fig:baseresults} where $M\in[3,50]$ and $\bar N\in[1,8]$. 
It should be noted that for $M=1$, a single full joint probability over all features is modelled while for $\bar N = 1$ the RFe corresponds to a Naive Bayes classifier. 
In general, an increase in either of both parameters leads to a better performance. 
More ferns (i.e. larger $M$) mean that the ensemble consists of more weak learners that contribute to the final estimate. 
Larger ferns (i.e. larger $\bar N$) mean using more binary features, thus increasing the probability to extract descriptive properties and decreasing the overlap of the feature distribution of different classes. 
Interestingly, while the performance increase saturates for both parameters it does not decrease, i.e. at least within the selected parameter range, the RFe does not overfit with increasing fern number or size. 
For the smallest RFe, average accuracy is with $28\%$ barely larger than random guessing. 
The best average accuracy of $70\%$ is achieved with the largest RFe. 
The confusion matrix in Table~\ref{tab:confusion_matrix_baseline} shows that the city and water classes are recognized best ($>80\%$ accuracy), followed by the field and forest classes ($>70\%$ accuracy). 
The largest confusion in terms of both, false negatives and false positives, is present for the street class which is only recognized to $38.5\%$. 
Figure~\ref{fig:base-res} shows that in general the proposed RFe classifier is able to obtain results that are highly consistent with the reference data. 
It should be stressed that (similar to the Random Forests in~\cite{hansch2018skipping}) no preprocessing (e.g. speckle reduction) or explicit feature extraction (apart from computing the polarimetric covariance matrices) is performed but the proposed RFe classifier works directly on the PolSAR data.

Increasing either fern number or fern size leads to an increased computational load as more binary features have to be evaluated. Figure~\ref{fig:time_contour_fernNumberSize} shows the amount of time required to train RFe of a certain size. 
The results for prediction time are similar (just scaled by a different number of samples) and omitted here for brevity. 
The run time complexity of training/applying a RFe is $\mathcal{O}(|D|\cdot N)$, where $|D|$ denotes the number of samples and $N$ the number of binary features.

\begin{table}[ht]
    \scriptsize
    \centering
    \begin{tabular}{c||c|c|c|c|c}
\hline \multicolumn{6}{c}{OA = 73.6$\pm$1.0, AA = 69.7$\pm$0.6, $\kappa$ = 52.3$\pm$1.4, F1 = 65.6$\pm$1.0, mIoU = 50.5$\pm$1.3} \\
\hline & \cellcolor[HTML]{FF0000}City & \cellcolor[HTML]{FFFF00}Field & \cellcolor[HTML]{008000}Forest & \cellcolor[HTML]{0000FF}Water & \cellcolor[HTML]{FF00FF}Street \\
\hline
\hline City & $ \textbf{80.1 (0.3)} $ & $ 0.6 \pm 0$ & $ 11.1 \pm 0.4$ & $ 0.2 \pm 0.1$ & $ 8.0 \pm 0.2$ \\
 \hline Field  & $ 1.0 \pm 0$ & $ \textbf{73.1 $ \pm $ 1.2}$ & $ 13.2 \pm 0.8$ & $ 0.6  \pm 0.2$ & $ 12.1 \pm 0.3$ \\
\hline Forest & $ 3.8 \pm 0.2$ & $ 10.2\pm 0.5$  & $ \textbf{72.4 $\pm$ 0.7}$ & $ 0.8 \pm 0.1$ & $ 12.8 \pm 0.1$ \\
 \hline Water & $ 8 \pm 0.9$ & $ 0.4 \pm 0.1$ & $ 2.4 \pm 0$ & $ \textbf{84.3 $\pm$ 0.3}$ & $ 4.8 \pm 0.5$ \\
\hline Street & $ 6.3 \pm 0.3$ & $ 38.5 \pm 0.6$ & $ 16.1 \pm 0.5$ & $ 0.6\pm 0.2$ & $ \textbf{38.5 $\pm$ 1.1}$ \\
\hline
\end{tabular}
    \caption{Confusion matrix of the RFe without any optimization ($M=30, \bar N=8$).}
    \label{tab:confusion_matrix_baseline}
\end{table}

\subsubsection{RFe optimization}
\begin{figure}
\begin{minipage}[t]{0.24\textwidth}
\centering
\includegraphics[width=\textwidth]{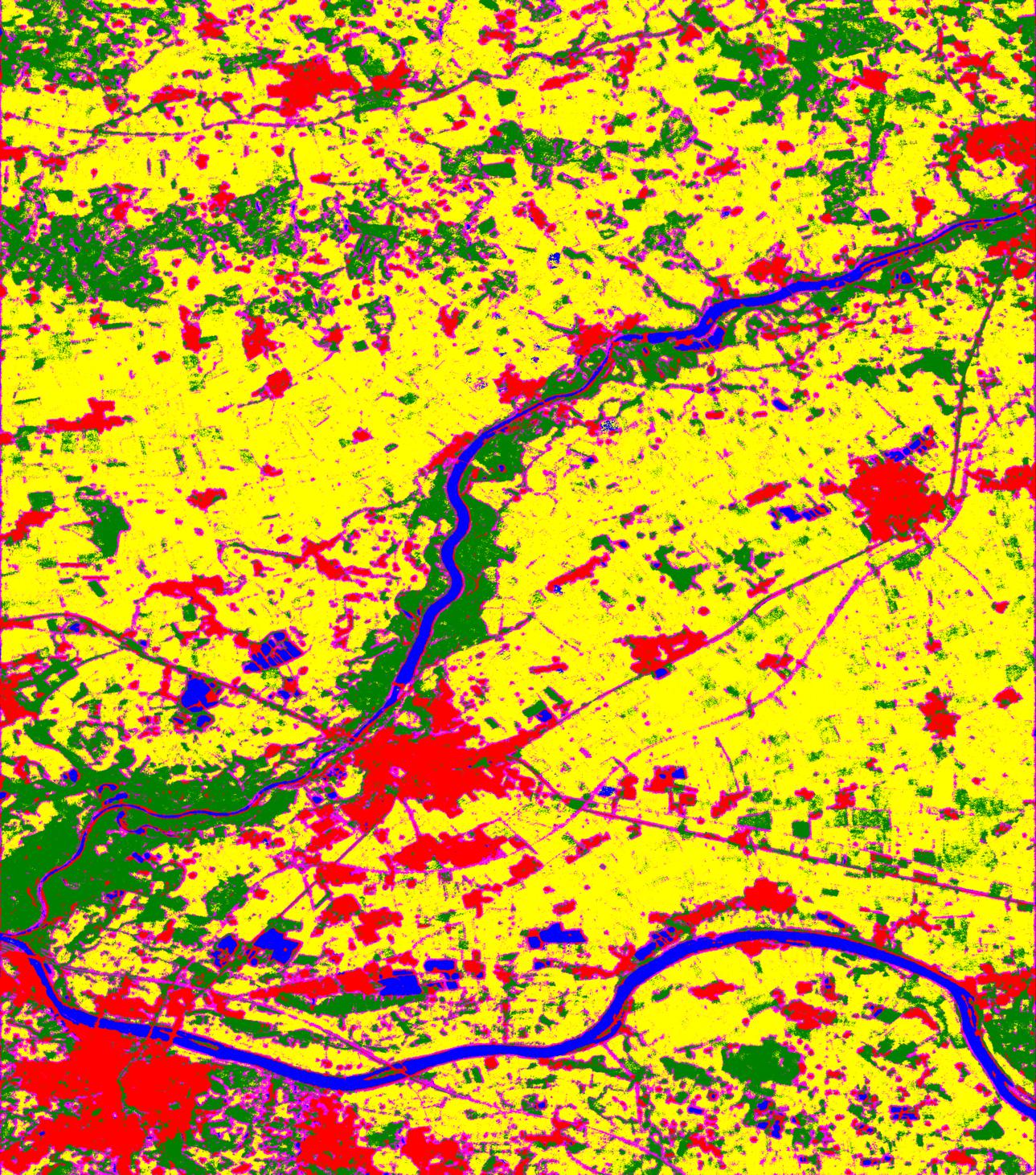}
\captionsetup{justification=centering}
\subcaption{Pre-selection and grouping ($M=30, \bar N=8$).}
\label{fig:quantRes-presel}
\end{minipage}
\begin{minipage}[t]{0.24\textwidth}
\centering
\includegraphics[width=\textwidth]{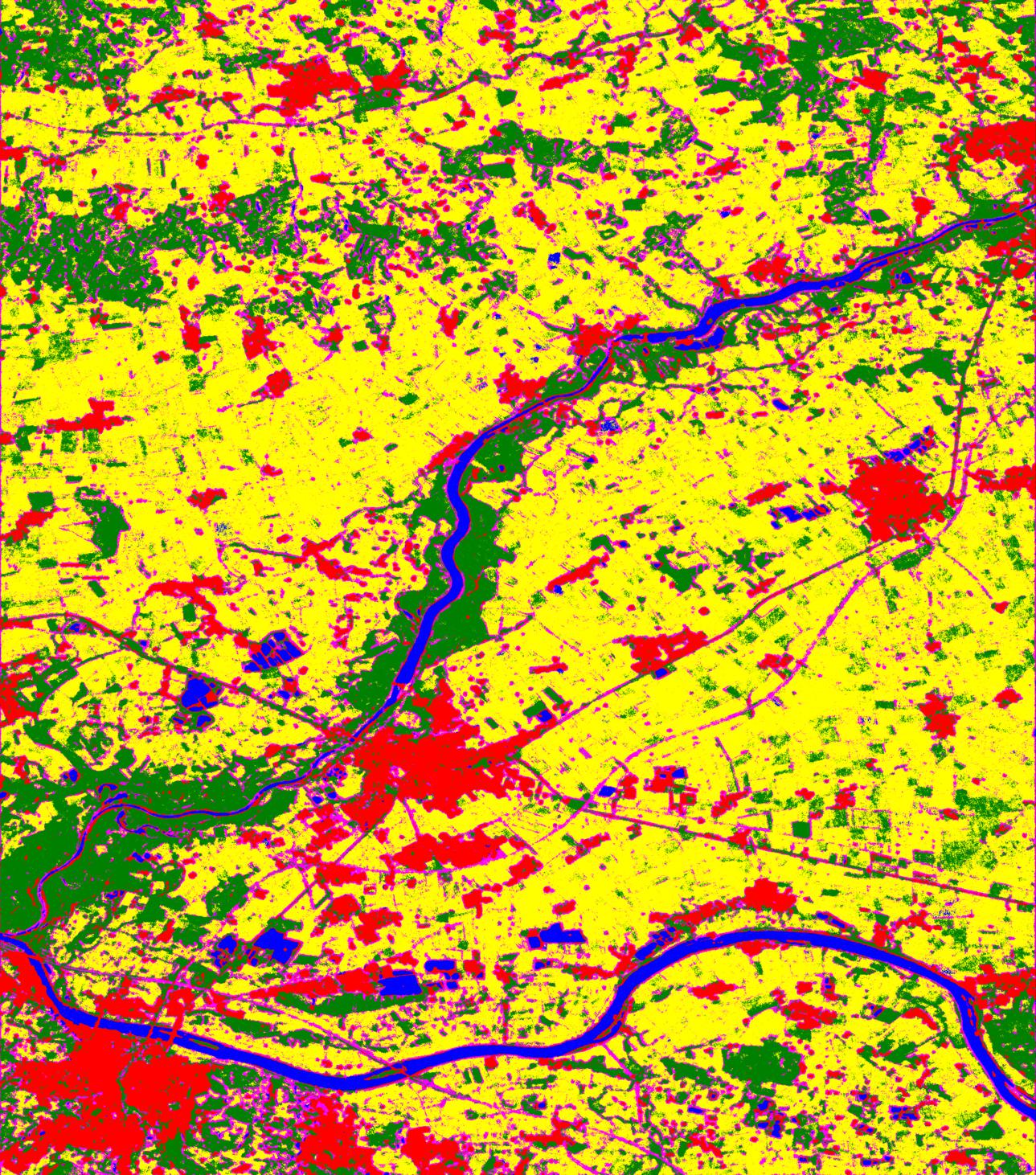}
\captionsetup{justification=centering}
\subcaption{Iterative optimization ($M=31, \bar N=6.3$ with $N_j\in[4,10]$).}
\label{fig:quantRes-iter}
\end{minipage} 
  \caption{Semantic maps obtained by RFe with different optimization strategies.}
  \label{fig:timecomplexity}
\end{figure}

Applying the feature selection and grouping step described in Section~\ref{sec:presel} increases the accuracy for all classes by at least $\sim 3\%$ as shown in Table~\ref{tab:confusion_matrix_feature_selection}. 
The largest improvement is achieved for the street class, which is now correctly detected to $44\%$. 
Consequently, the average accuracy is raised from $70\%$ (using no optimization) to $74\%$.
The obtained semantic map (shown in Figure~\ref{fig:quantRes-presel}) is significantly smoother. 
We use the same parameter setting as above (Sec.~\ref{sec:baselineRes}) with $M=30, \bar N=8$. 
The lower bound on the information gain (Eq.~\ref{eq:infogain}) is set to $0.01$ and the upper bound on the feature correlation (Eq.~\ref{eq:corr}) to $0.9$. 
Using stricter bounds, in particular on the information gain, would lead to potentially stronger features. However, it would also increase the number of features to be tested and thus increase the computational load. With the used setting, feature selection, grouping, and fern training took $20-23sec$, i.e. approximately a factor three longer than training the baseline. 
As the number and size of the ferns is fixed for this optimization scheme (and was selected to be identical to the baseline), the prediction time does not change.

\begin{table}[ht]
\scriptsize
\begin{minipage}[t]{\columnwidth}
    \centering
    \begin{tabular}{c||c|c|c|c|c}
\hline \multicolumn{6}{c}{OA = 77.8$\pm$0.2, AA = 73.6$\pm$0.0, $\kappa$ = 58.5$\pm$0.1, F1 = 70.0$\pm$0.1, mIoU = 56.2$\pm$0.1} \\
\hline & \cellcolor[HTML]{FF0000}City & \cellcolor[HTML]{FFFF00}Field & \cellcolor[HTML]{008000}Forest & \cellcolor[HTML]{0000FF}Water & \cellcolor[HTML]{FF00FF}Street \\
\hline
\hline City & $ \textbf{83.9 $\pm$ 0.6} $ & $ 0.3 \pm 0$ & $ 8.5 \pm 0.5$ & $ 0.1 \pm 0$ & $ 7.3 \pm 0.1$ \\
 \hline Field  & $ 0.9 \pm 0$ & $ \textbf{77.7 $ \pm $ 0.4}$ & $ 9.1 \pm 0$ & $ 0.2  \pm 0$ & $ 12.1 \pm 0.4$ \\
\hline Forest & $ 3.6 \pm 0.2$ & $ 7.6\pm 0.8$  & $ \textbf{75.0 $\pm$ 0.6}$ & $ 0.7 \pm 0$ & $ 13.2 \pm 0$ \\
 \hline Water & $ 5.9 \pm 0.5$ & $ 0.2 \pm 0$ & $ 2.2 \pm 0$ & $ \textbf{87.4 $\pm$ 0.5}$ & $ 4.3 \pm 0.1$ \\
\hline Street & $ 6.0 \pm 0.2$ & $ 36.4 \pm 0.1$ & $ 13.2 \pm 0$ & $ 0.2\pm 0$ & $ \textbf{44.2 $\pm$ 0.1}$ \\
\hline
\end{tabular}
        \subcaption{Confusion matrix applying feature selection and grouping ($M=30, \bar N=8$)}
    \label{tab:confusion_matrix_feature_selection}
\end{minipage}
\begin{minipage}[t]{\columnwidth}
    \begin{tabular}{c||c|c|c|c|c}
\hline \multicolumn{6}{c}{OA = 77.2$\pm$0.5, AA = 73.8$\pm$0.3, $\kappa$ = 57.7$\pm$0.7, F1 = 69.6$\pm$0.5, mIoU = 55.4$\pm$0.6} \\
\hline & \cellcolor[HTML]{FF0000}City & \cellcolor[HTML]{FFFF00}Field & \cellcolor[HTML]{008000}Forest & \cellcolor[HTML]{0000FF}Water & \cellcolor[HTML]{FF00FF}Street \\
\hline
\hline City & $ \textbf{84.2 $\pm$ 0.3} $ & $ 0.3 \pm 0$ & $ 8.5 \pm 0.4$ & $ 0.1 \pm 0$ & $ 6.9 \pm 0.1$ \\
 \hline Field  & $ 1.0 \pm 0$ & $ \textbf{76.5 $ \pm $ 0.7}$ & $ 10.6\pm 0.5$ & $ 0.3  \pm 0.1$ & $ 11.6 \pm 0.1$ \\
\hline Forest & $ 3.7 \pm 0$ & $ 6.7\pm 0.1$  & $ \textbf{76.2 $\pm$ 0.1}$ & $ 0.7 \pm 0$ & $ 12.8 \pm 0$ \\
 \hline Water & $ 5.6 \pm 0.1$ & $ 0.2 \pm 0$ & $ 2.1 \pm 0.1$ & $ \textbf{87.6 $\pm$ 0}$ & $ 4.6 \pm 0$ \\
\hline Street & $ 6.1 \pm 0.1$ & $ 35.4 \pm 0.1$ & $ 13.6 \pm 0.5$ & $ 0.3\pm 0.1$ & $ \textbf{44.7 $\pm$ 0.3}$ \\
\hline
\end{tabular}
        \subcaption{Confusion matrix applying iterative fern optimization (on average, $M=27.9\pm3.2$, $\bar N = 6.3\pm0.2$)}
    \label{tab:confusion_matrix_iterative}
\end{minipage}
\begin{minipage}[t]{\columnwidth}
    \begin{tabular}{c||c|c|c|c|c}
\hline \multicolumn{6}{c}{OA = 77.6$\pm$0.3, AA = 73.5$\pm$0.3, $\kappa$ = 58.3$\pm$0.3, F1 = 69.4$\pm$0.3, mIoU = 55.5$\pm$0.3} \\
\hline & \cellcolor[HTML]{FF0000}City & \cellcolor[HTML]{FFFF00}Field & \cellcolor[HTML]{008000}Forest & \cellcolor[HTML]{0000FF}Water & \cellcolor[HTML]{FF00FF}Street \\
\hline
\hline City & $ \textbf{84.2 $\pm$ 0.2} $ & $ 0.3 \pm 0$ & $ 8.4 \pm 0$ & $ 0.1 \pm 0$ & $ 6.9 \pm 0.2$ \\
 \hline Field  & $ 1.0 \pm 0$ & $ \textbf{77.2 $ \pm $ 0.5}$ & $ 10.3\pm 0.4$ & $ 0.3  \pm 0$ & $ 11.2 \pm 0.1$ \\
\hline Forest & $ 3.9 \pm 0.2$ & $ 6.9\pm 0.3$  & $ \textbf{75.9 $\pm$ 0.8}$ & $ 0.7 \pm 0$ & $ 12.6 \pm 0.3$ \\
 \hline Water & $ 5.8 \pm 0.1$ & $ 0.2 \pm 0$ & $ 2.1 \pm 0.1$ & $ \textbf{87.5 $\pm$ 0.5}$ & $ 4.4 \pm 0.3$ \\
\hline Street & $ 6.3 \pm 0$ & $ 36.9 \pm 1.1$ & $ 13.7 \pm 0.5$ & $ 0.2\pm 0$ & $ \textbf{42.8 $\pm$ 1.6}$ \\
\hline
\end{tabular}
        \subcaption{Confusion matrix applying two fern optimizations}
    \label{tab:confusion_matrix_two_optimizations}
\end{minipage}
\caption{Confusion matrices obtained by RFe with different optimization strategies.}
\end{table}

To evaluate the second proposed optimization scheme, we initialize a RFe with five ferns of size six. 
As described in Section~\ref{sec:itopt}, random changes are applied to the current model. Only if a change improves model performance it is kept, otherwise it is rejected. 
The optimization is repeated for at least $IT_{min}$ iterations and terminated if no improvement is achieved after $\Delta$ consecutive iterations. 
In the following experiments, we used $IT_{min}=30$ and $\Delta=15$. The other parameters remain as in Section~\ref{sec:baselineRes}. 

\begin{figure}[!htb]
\begin{minipage}[t]{0.24\textwidth}
\centering
\includegraphics[width=\textwidth]{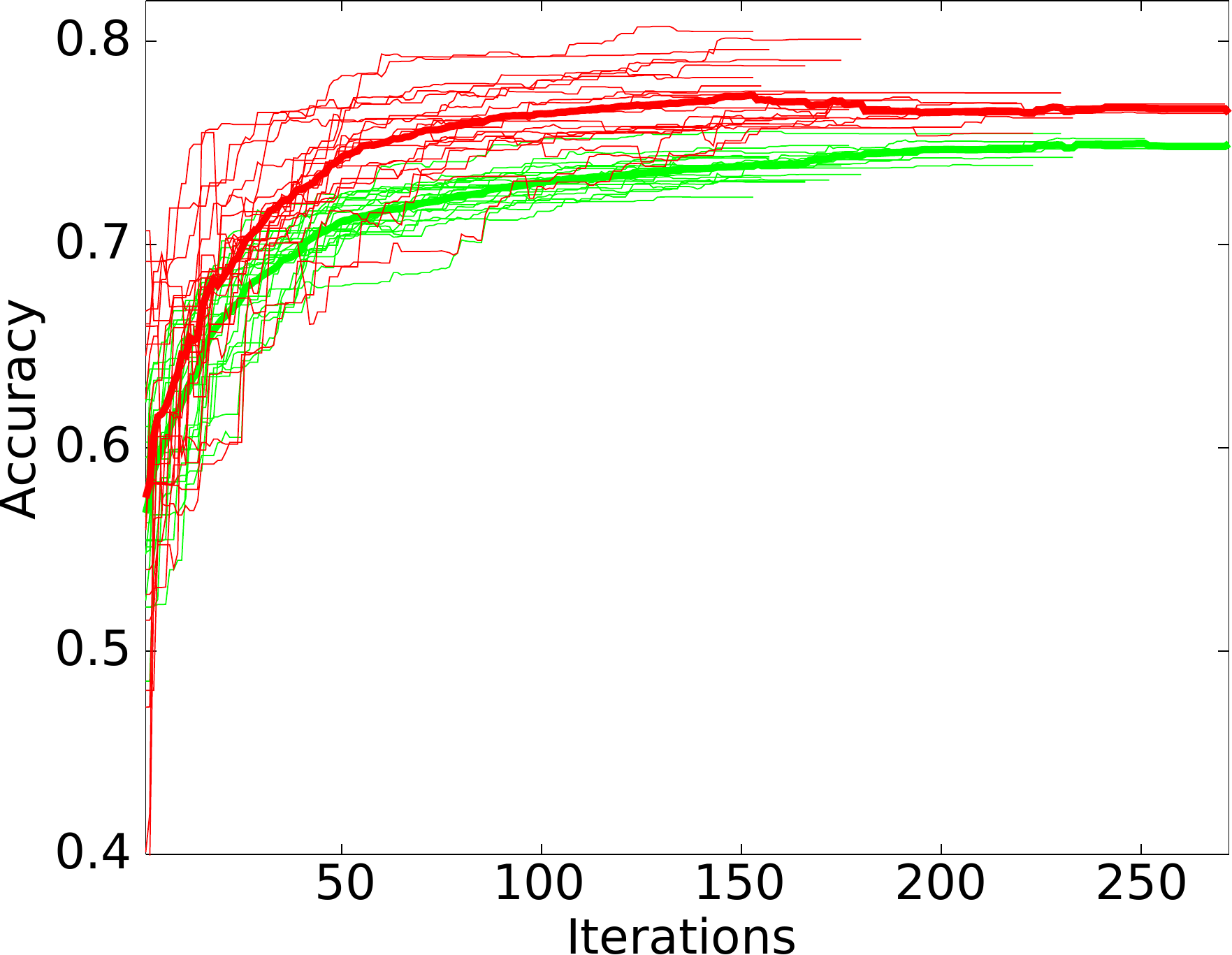}
\captionsetup{justification=centering}
\subcaption{Overall accuracy on train (green) and test (red) data.}
\label{subfig:accuracy}
\end{minipage}
\begin{minipage}[t]{0.24\textwidth}
\centering
\includegraphics[width=\textwidth]{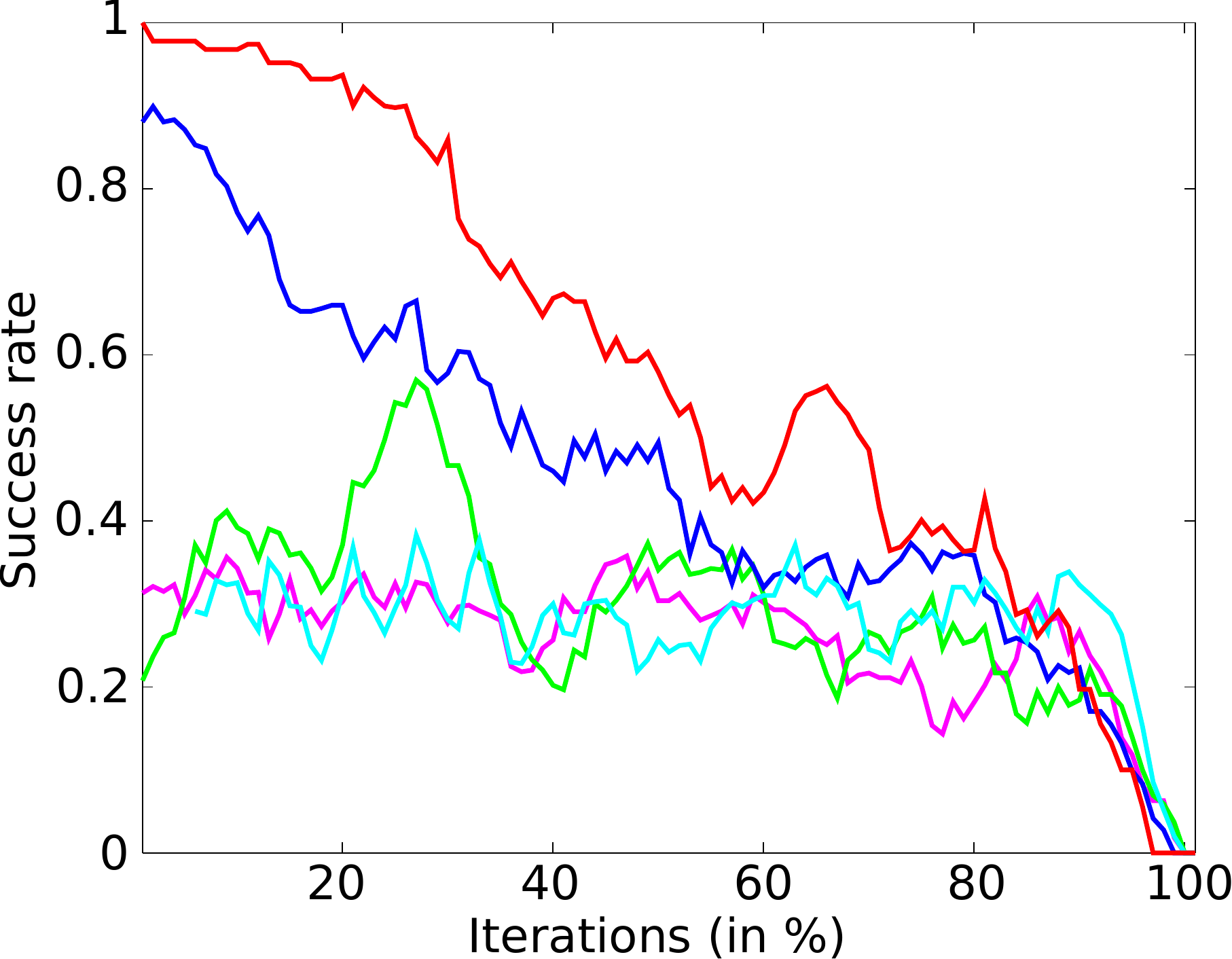}
\captionsetup{justification=centering}
\subcaption{Success rate of different changes: Add a fern (red), add (blue) or delete (cyan) a feature, switch two features (green), adjust split threshold (magenta).}
\label{subfig:changes}
\end{minipage}
\begin{minipage}[t]{0.24\textwidth}
\centering
\includegraphics[width=\textwidth]{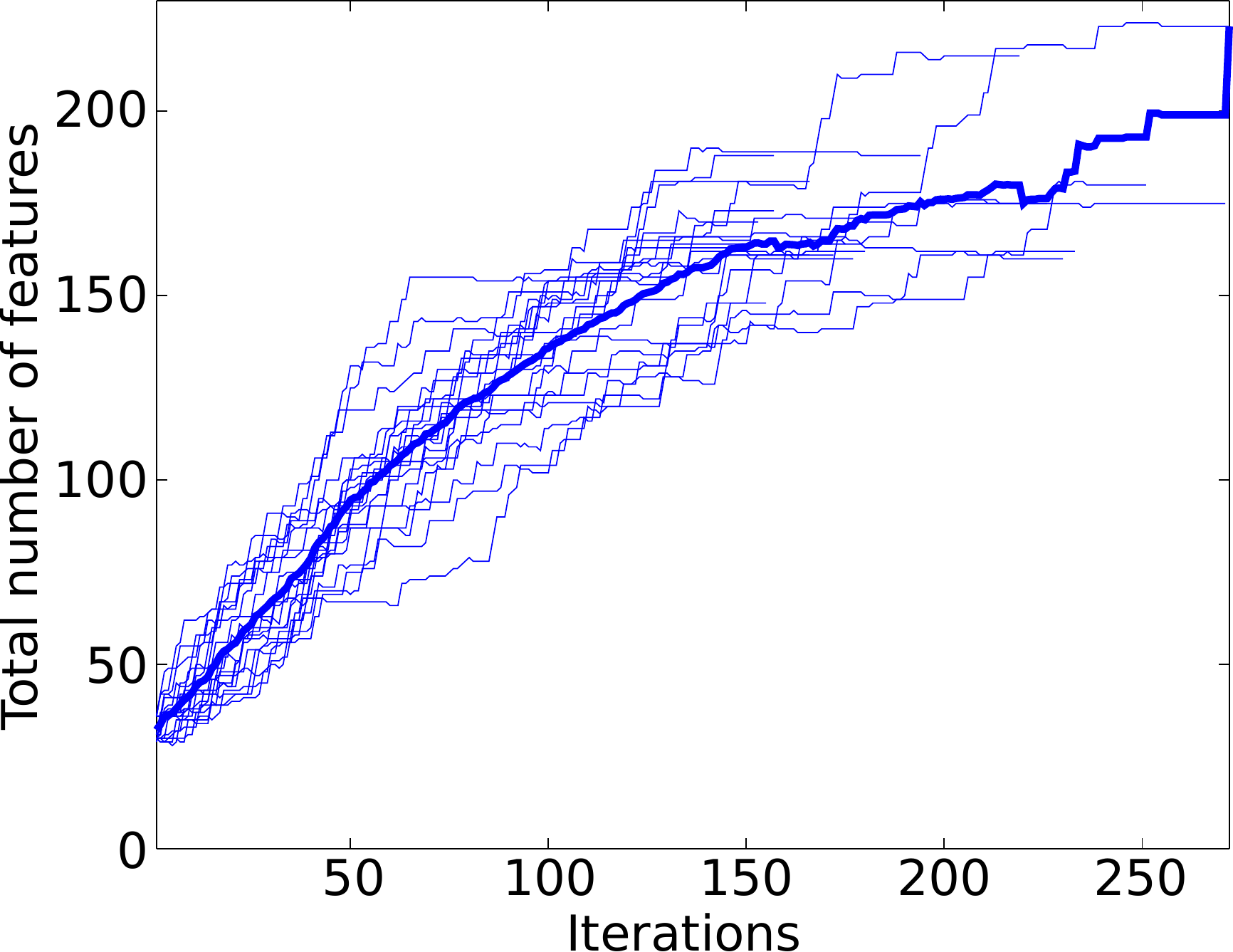}
\captionsetup{justification=centering}
\subcaption{Number of features.}
\label{subfig:feat}
\end{minipage}
\begin{minipage}[t]{0.24\textwidth}
\centering
\includegraphics[width=\textwidth]{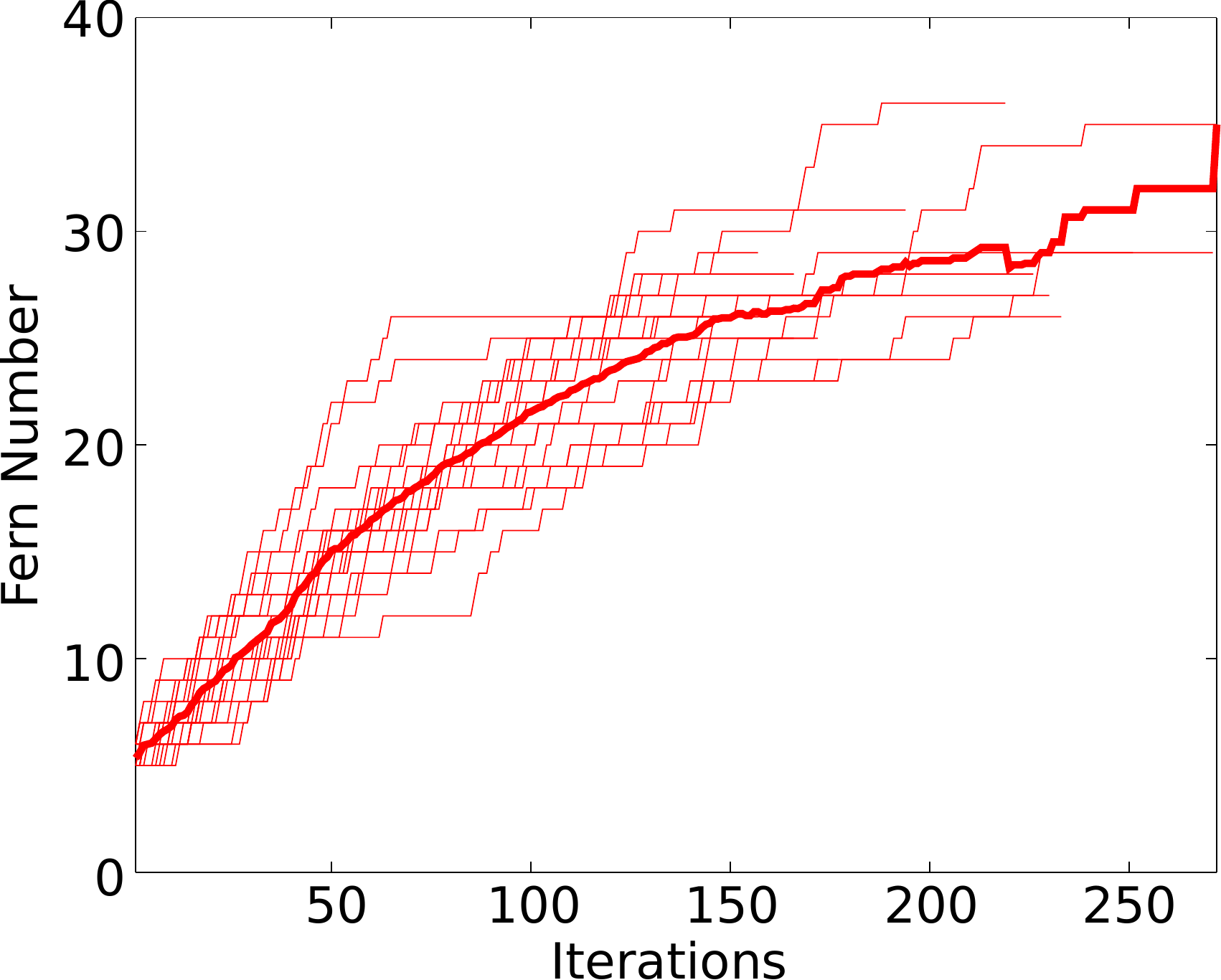}
\captionsetup{justification=centering}
\subcaption{Number of feature groups.}
\label{subfig:featg}
\end{minipage}
    \caption{Detailed changes of fern properties over optimization iterations in different folds and runs.}
    \label{fig:fernProperties}
\end{figure}

Figure~\ref{fig:fernProperties} illustrates the changes over the iterations for the different RFe for four repetitions over five folds (i.e. $20$ runs in total) as well as the overall average. Note, that the optimization in different runs terminates after different numbers of iterations. The average for a given iteration is only computed over runs with at least that many iterations. 
The initial RFe has five ferns with six features each and reaches an accuracy of $AA = 57.5\% \pm 9\%$. 
As can be seen in Figure~\ref{subfig:accuracy}, training as well as test accuracy increases and saturates before optimization is automatically terminated. 
On average, the final RFe classifiers have $M=27.9\pm3.2$ ferns with $\bar N = 6.3\pm0.2$ features and reach an accuracy of $AA=77.5\%\pm1.4\%$. 
While all possible changes (see Section~\ref{sec:itopt}) have equal chance to be applied at any given iteration, Figure~\ref{subfig:changes} shows that adding features - either a single feature within a group or a whole new group, i.e. six new features at once - have a much higher acceptance probability at the beginning of the optimization process. 
Consequently, the number of features as well as the number of feature groups grows. 
The increase in the number of features is dominated by adding more feature groups instead of increasing the number of features within the existing groups (see Figures~\ref{subfig:feat} and~\ref{subfig:featg}). 
Deleting an existing feature, switching two features among feature groups, or changing the split threshold have a roughly equal success rate over the whole optimization process.

The iteratively optimized RFe achieves (on average) quantitative results (see Table~\ref{tab:confusion_matrix_iterative}) as well as qualitative results (see Figure~\ref{fig:quantRes-iter}) that are very similar to the preselection and grouping approach. 
The observed differences in the computed performance metrics are neither consistent nor significant. 
However, it appears that the variance of the results is slightly larger. 
It should be noted, however, that preselection and grouping requires defining two parameters that are difficult to set, while the iterative procedure only depends on the termination criteria which only marginally influences the final performance.

The training time increased to $30-110sec$, i.e. up to five times more than for the preselection and grouping approach. 
Since the obtained number of features (and feature groups) is similar to the manually set parameters of the baseline, the application time does not change significantly. 

Preselection and grouping evaluates features only either individually (for the preselection) or in pairs (for the grouping). 
Thus, it might be reasonable to use this approach to generate an initial fern that is then further optimized by the iterative training strategy. 
The results in Table~\ref{tab:confusion_matrix_two_optimizations} show that this does not change the performance significantly. 
This indicates that, at least for the given dataset, higher order relations between features are well modelled in the feature groups generated by the preselection and grouping approach despite its greedy nature. 

\begin{figure}[!htb]
\begin{minipage}[t]{0.24\textwidth}
\centering
\includegraphics[width=\textwidth]{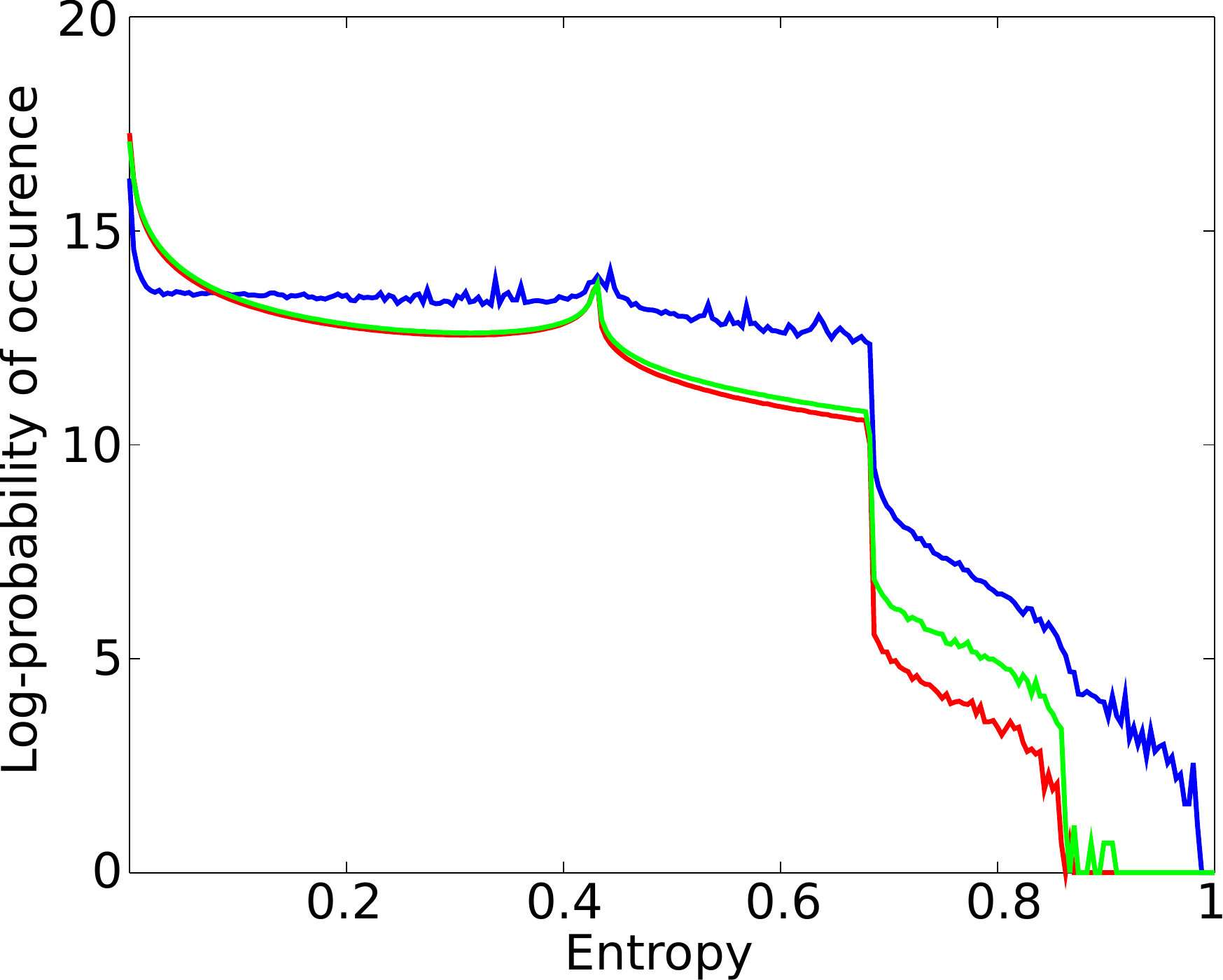}
\captionsetup{justification=centering}
\subcaption{Log-probability of posterior entropy.}
\label{subfig:entropy}
\end{minipage}
\begin{minipage}[t]{0.24\textwidth}
\centering
\includegraphics[width=\textwidth]{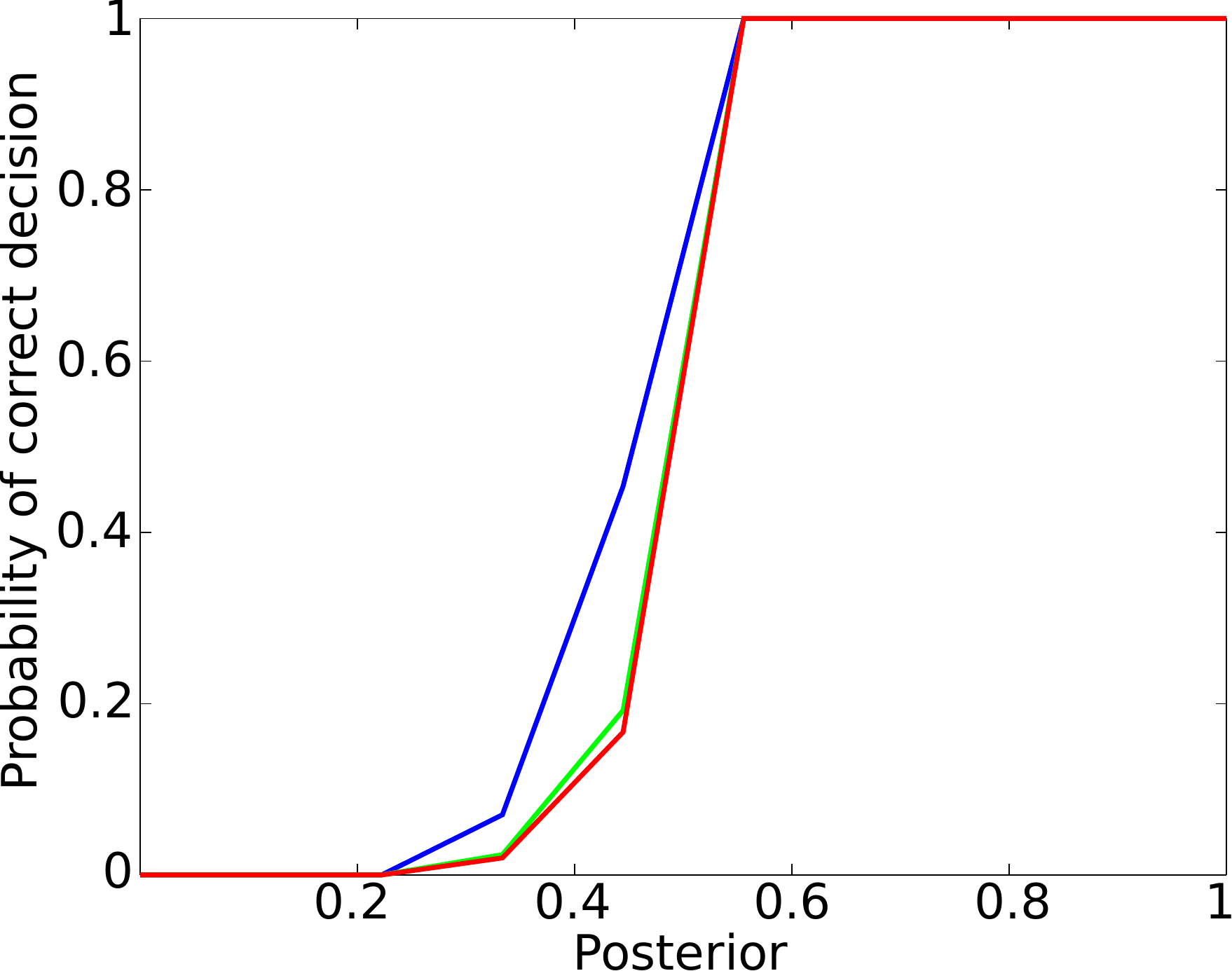}
\captionsetup{justification=centering}
\subcaption{Calibration curves.}
\label{subfig:calib}
\end{minipage}
    \caption{Classifier uncertainty and calibration for the baseline RFe (blue), pre-selection and grouping (green), and iterative optimization (red).}
    \label{fig:entropy}
\end{figure}

RFe do not only provide an estimate of the class label but instead estimate the full class posterior probability. 
This allows judging the certainty of the classifier in its decision, e.g. based on the entropy~$H$ of the class posterior: A low uncertainty corresponds to $H=0$ which is achieved if one class obtained $100\%$ of the probability mass. A high uncertainty corresponds to $H=1$ which means that all classes were assigned equal probability. 
Figure~\ref{subfig:entropy} shows the histogram of entropy values for the RFe baseline as well as the two optimization techniques. 
The similarity in performance of both optimization strategies can also be observed here: 
Both achieve significantly more estimates with low uncertainty than the baseline. 
The iterative scheme is slightly superior to the preselection and grouping approach. 
Note that the small peak at $H=0.43$ visible in all three graphs corresponds to the situation where two classes are assigned $50\%$ probability (i.e. $H=-2\cdot0.5\log_5(0.5)$). 
The strong decrease at $H=0.68$ corresponds to the situation where three classes are assigned equal probability (i.e. $H=-3\cdot0.33\log_5(0.33)$), which happens significantly less often. 

A high certainty in the estimate does not necessarily warrant its correctness. 
Figure~\ref{subfig:calib} shows that the RFe exhibits rather typical calibration curves~\cite{broecker2007}. 
It is underconfident for large posterior values (e.g. in all cases where the posterior of the dominant class was estimated as $70\%$, the decision was actually correct) and overconfident for small posterior values.

\subsubsection{Comparative Evaluation}
In this last section we evaluate the performance of the proposed RFe (with its different optimization strategies) with two other classifiers: A Random Forest and a deep neural network. Both are trained and evaluated in the same way as the RFe classifier, i.e. with 5-fold cross validation based on vertical image stripes of the Plattling dataset. 
For better comparisons, all experiments in this section are conducted by training with all available training samples. 

The Random Forest (RF) employs the same node projections (i.e. works directly on the PolSAR data as well). 
We choose two different parameter settings: First, a RF with 30 trees with a maximum height of eight (denoted as {\it RF-30-8}). Given the large number of samples, the maximum height is usually reached, i.e. each of the 30 base-learners (trees) evaluate roughly eight features per sample. 
This setting is most comparable with the RFe settings where 30 base-learners (ferns) evaluate (exactly) eight features. 
The difference is that while the RFe evaluates the same features for different samples, the RF applies potentially different binary tests to different samples. 
The second parameter setting creates a RF with ten trees with a maximal height of 30, i.e. less but higher trees (denoted as {\it RF-10-30}).

As neural network architecture we chose SegNet that was adapted in~\cite{Audebert2018} to remote sensing data. It consists of an encoder and decoder, both of them having five blocks of convolutional and pooling layers. 
A standard SGD optimizer is used with a learning rate of 0.01, weight decay of 0.005, and momentum of 0.9. 
The network is trained for ten epochs on all available training data. 
During training we apply flips of the image patches as data augmentation.

\begin{figure}
\begin{minipage}[t]{0.24\textwidth}
\centering
\includegraphics[width=\textwidth]{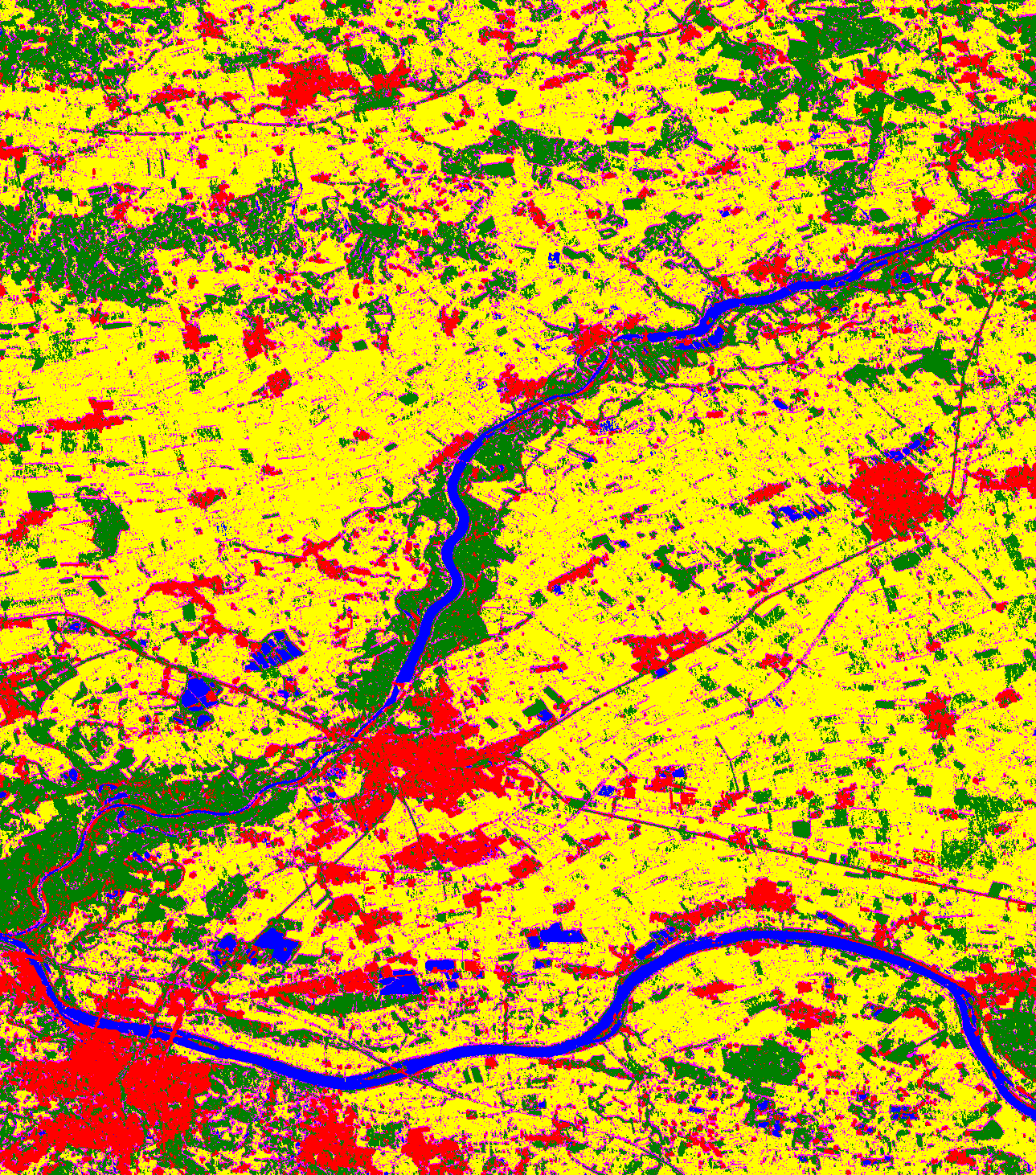}
\captionsetup{justification=centering}
\end{minipage}
\begin{minipage}[t]{0.24\textwidth}
\centering
\includegraphics[width=\textwidth]{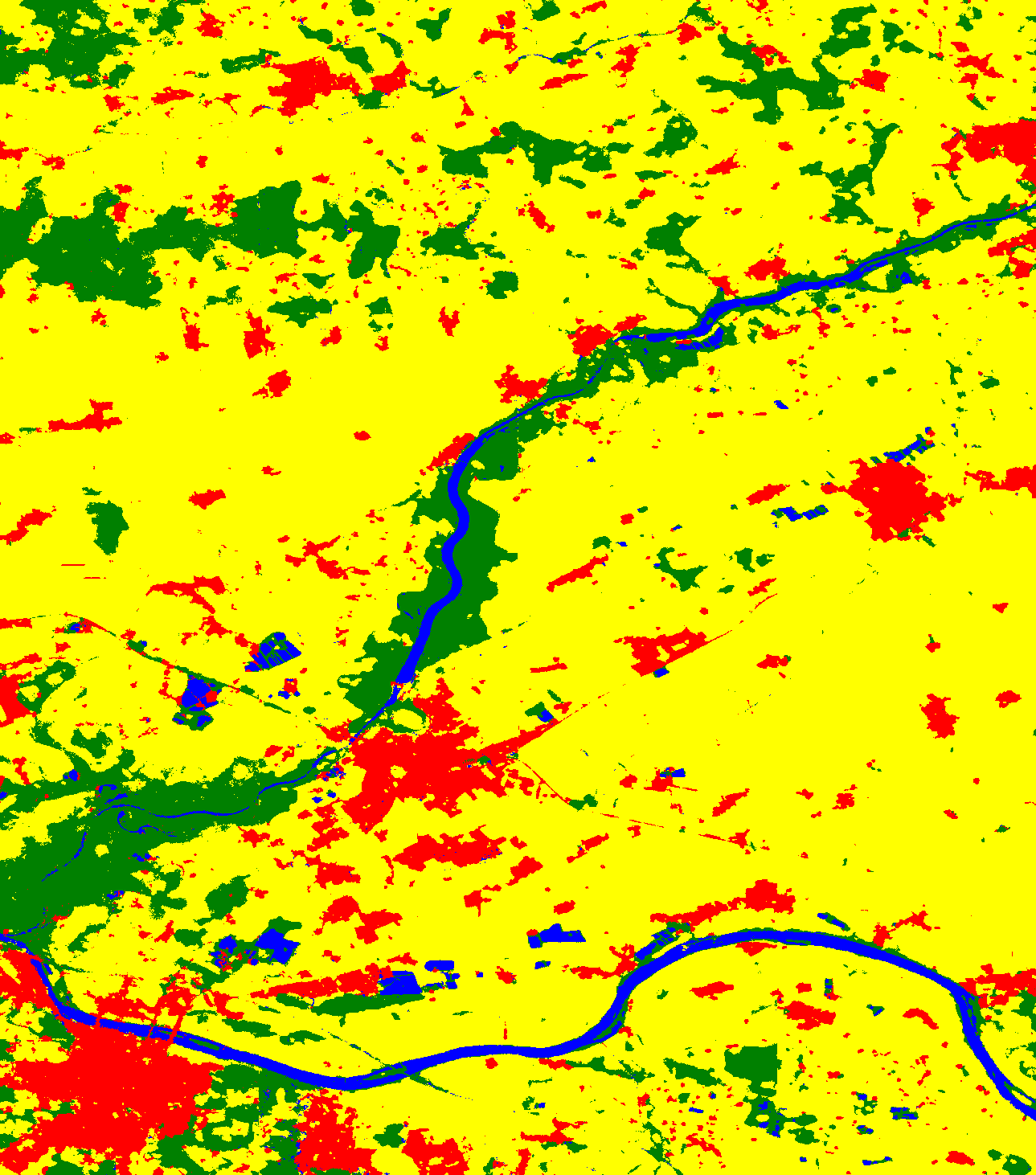}
\captionsetup{justification=centering}
\end{minipage} 
  \caption{Semantic maps obtained by a RF (i.e. {\it RF-30-8}) on the left and a deep network (i.e. SegNet) on the right.}
  \label{fig:qual-comp}
\end{figure}

\begin{figure}[!htb]
\centering
\includegraphics[width=\columnwidth]{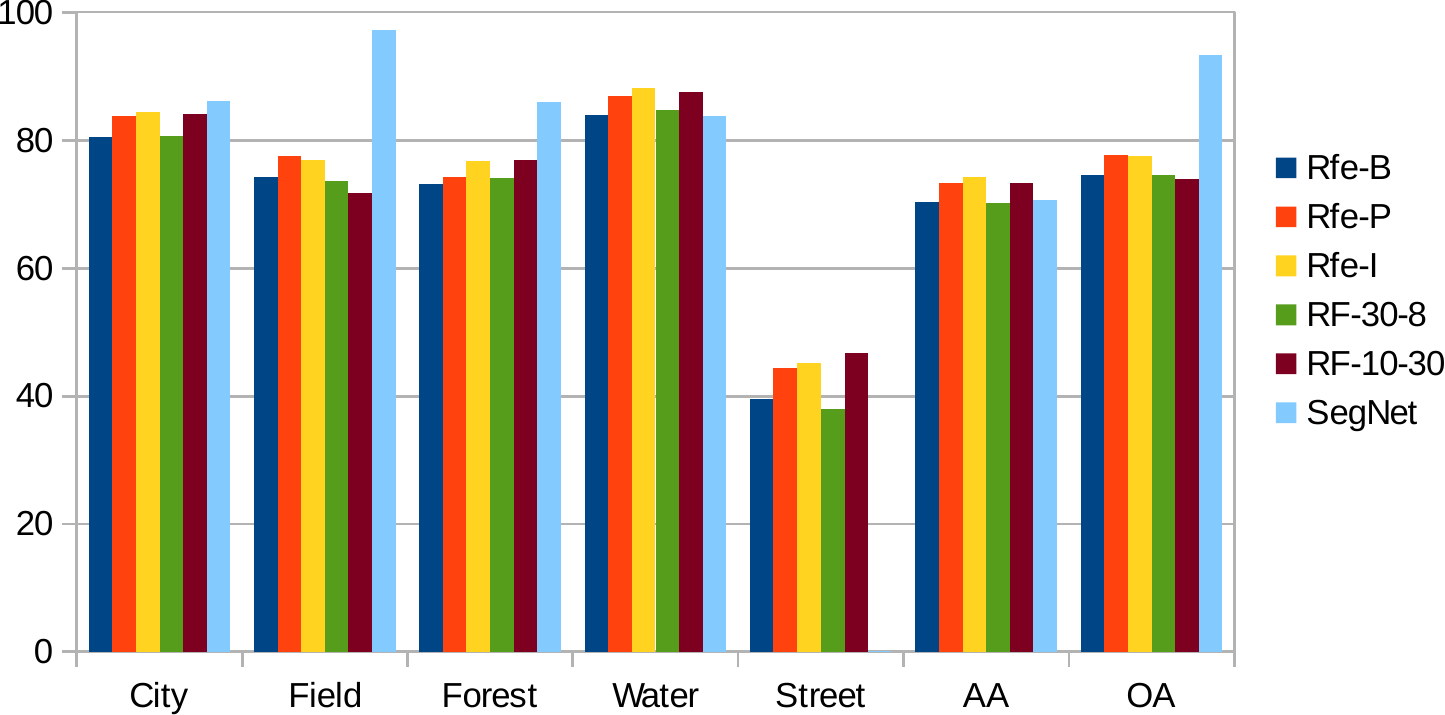}
\caption{Per-class, overall, and average accuracy of proposed RFe method, as well as a Random Forest, and SegNet.}
\label{fig:comp}
\end{figure}

The results in terms of per-class, average, and overall accuracy are shown in Figure~\ref{fig:comp}. 
Despite being a significantly simpler model, the RFe offers a competitive performance to the other approaches in particular in terms of average accuracy.  
As already discussed in the previous section, both RFe optimization schemes consistently outperform the baseline, while the difference between both optimization strategies is not significant. 

The large but shallow RF {\it RF-30-8} is on par with the RFe baseline (see also the left side of Figure~\ref{fig:qual-comp}).
Training and prediction times are $8\times$ and $1.6\times$ longer, respectively. 
The reason is the computational overhead produced by determining the path of a sample through the decision trees and applying different operations to different samples. This does not only require additional computations, but also prevents coalesced memory access. 
RFe extract the same binary features from all samples and thus does not perform the additional computations induced by the hierarchical structure of decision trees. 
Note, however, that for neither method any kind of parallelization is used. 
The small but high {\it RF-10-30} is mostly on par with the optimized RFe, with being slightly inferior for the field class but superior for the street class. 
It is more sensitive to minority classes that correspond to finer image structures. 
As a consequence, the overall accuracy of the RF is slightly worse than for the proposed RFe. 
Training and prediction times took more than $2\times$ longer than for {\it RF-30-8}. 
The results show that the proposed RFe is a valid alternative to RFs offering similar accuracy while being significantly easier to implement, to train, and to interpret.
Furthermore, it provides - even in a vanilla implementation - a considerable advantage with respect to computation time during training as well as during prediction.

The SegNet approach outperforms both, RFe and RF, for dominant classes partially by a large margin, e.g. $97.2\%$ in contrast to $71-78\%$ for the Field class. It is slightly better for the City class and slightly worse for the Water class (see the right side of Figure~\ref{fig:qual-comp} for qualitative results). 
The Road class is completely missed and mostly confused as Field (to $76\%$). 
Thus, while the overall accuracy is significantly larger than for RF and RFe, the average accuracy is lower. 
Training the SegNet needed 2.5h on a GeForce GTX 1060 GPU which is roughly half the time needed for the RFe approach. 
Test time is with 15min significantly faster than RF and RFe that needed roughly 100min. 
Note, however, that parallelization is neither used for RF nor for RFe but strongly exploited for SegNet. 

\begin{figure}[!htb]
\centering
\includegraphics[width=0.45\columnwidth]{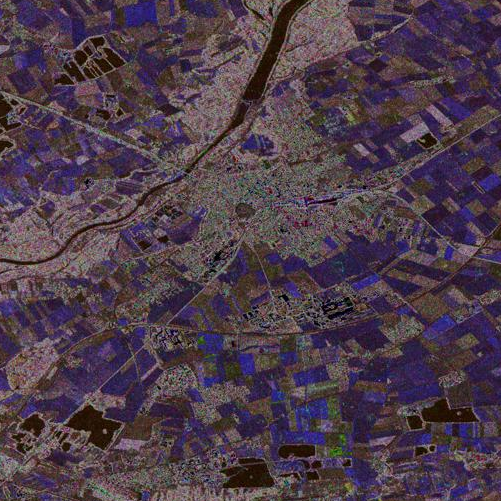}
\includegraphics[width=0.45\columnwidth]{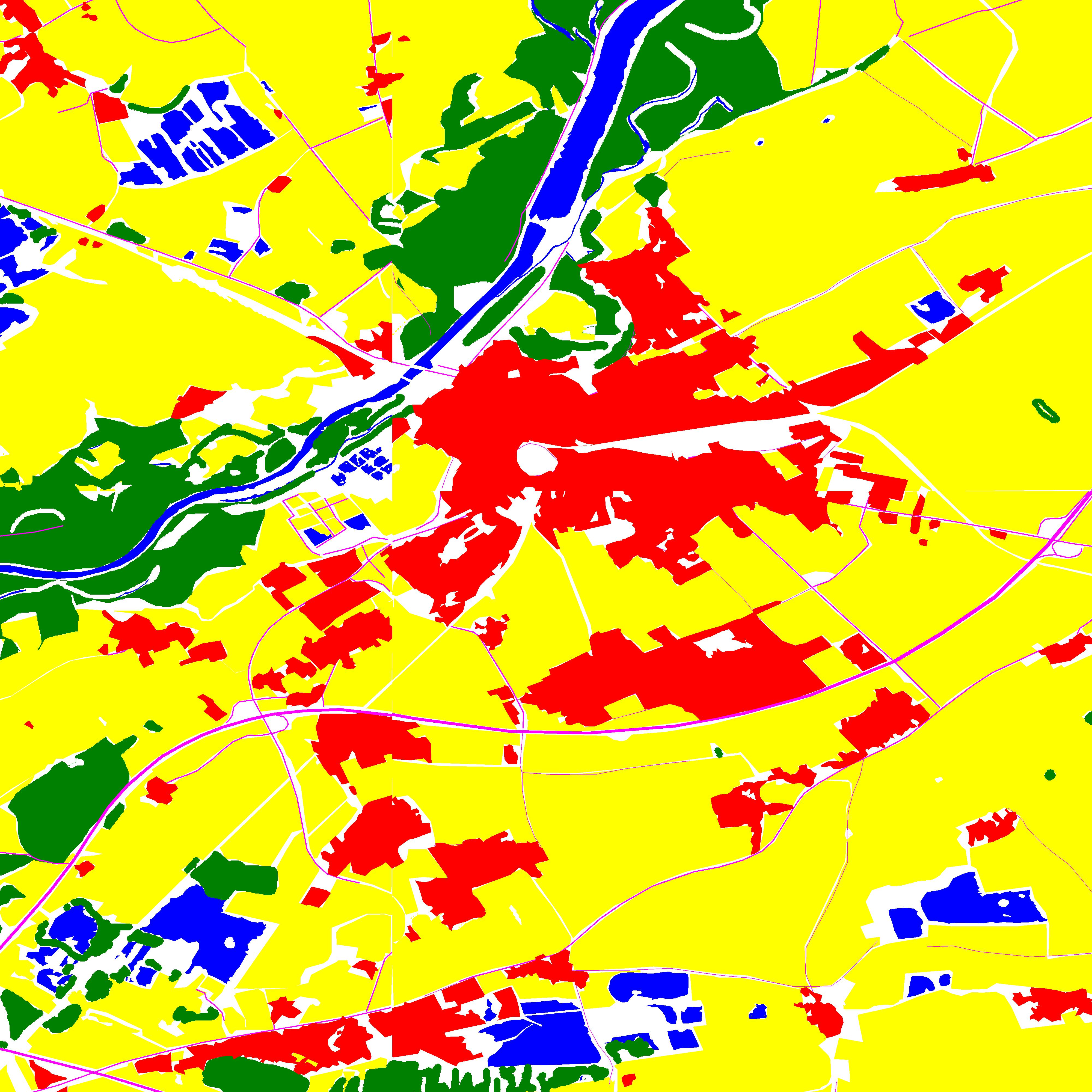}
\includegraphics[width=0.45\columnwidth]{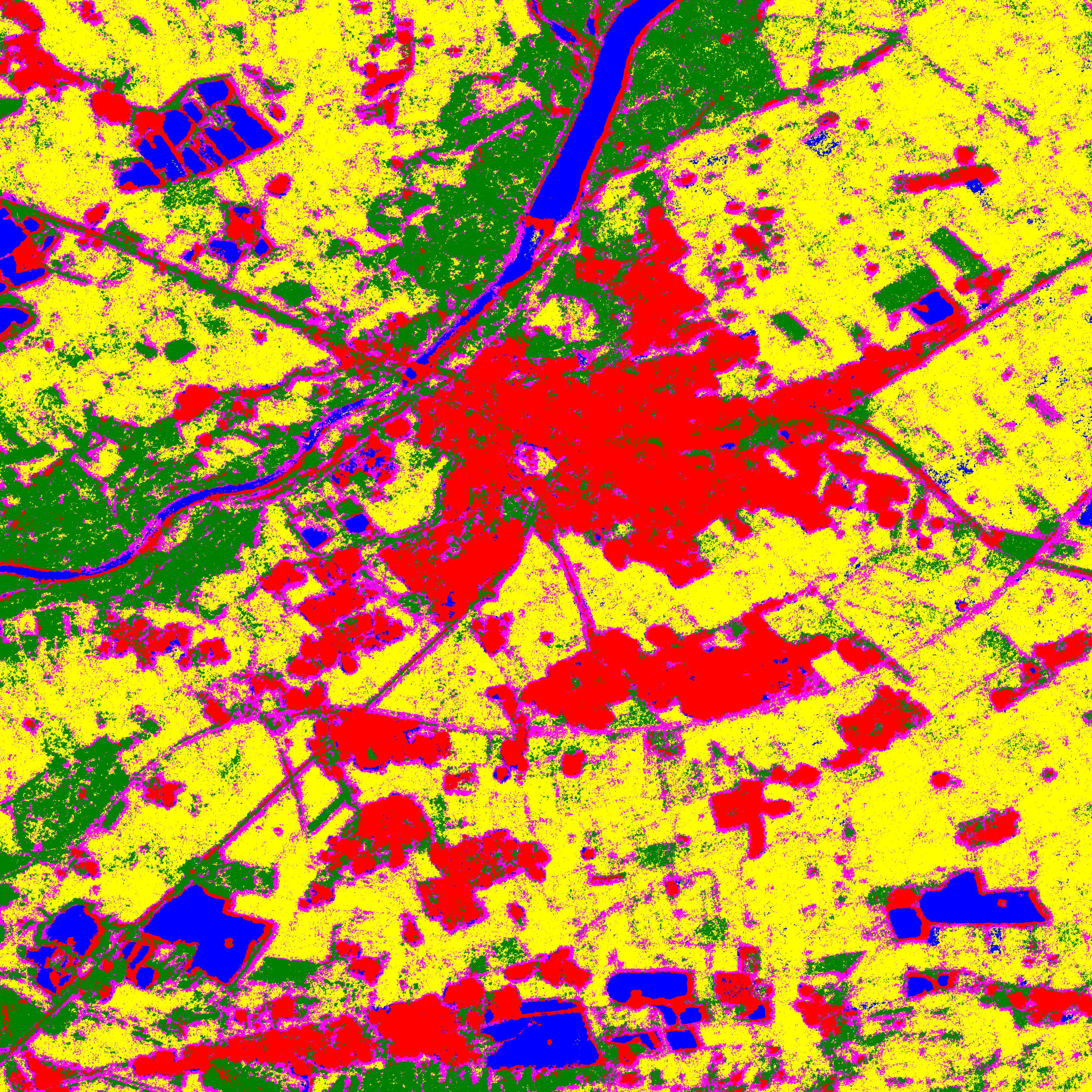}
\includegraphics[width=0.45\columnwidth]{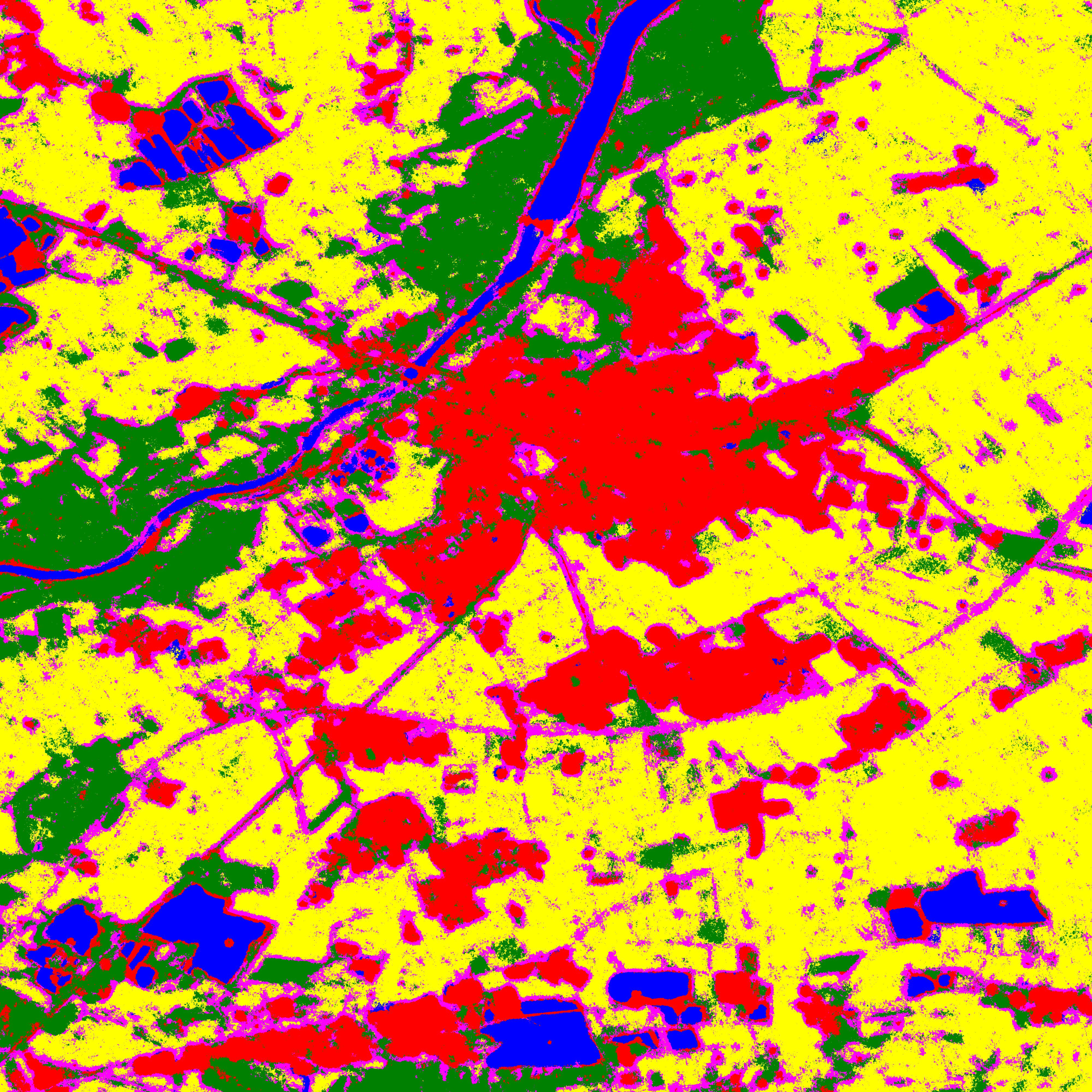}
\includegraphics[width=0.45\columnwidth]{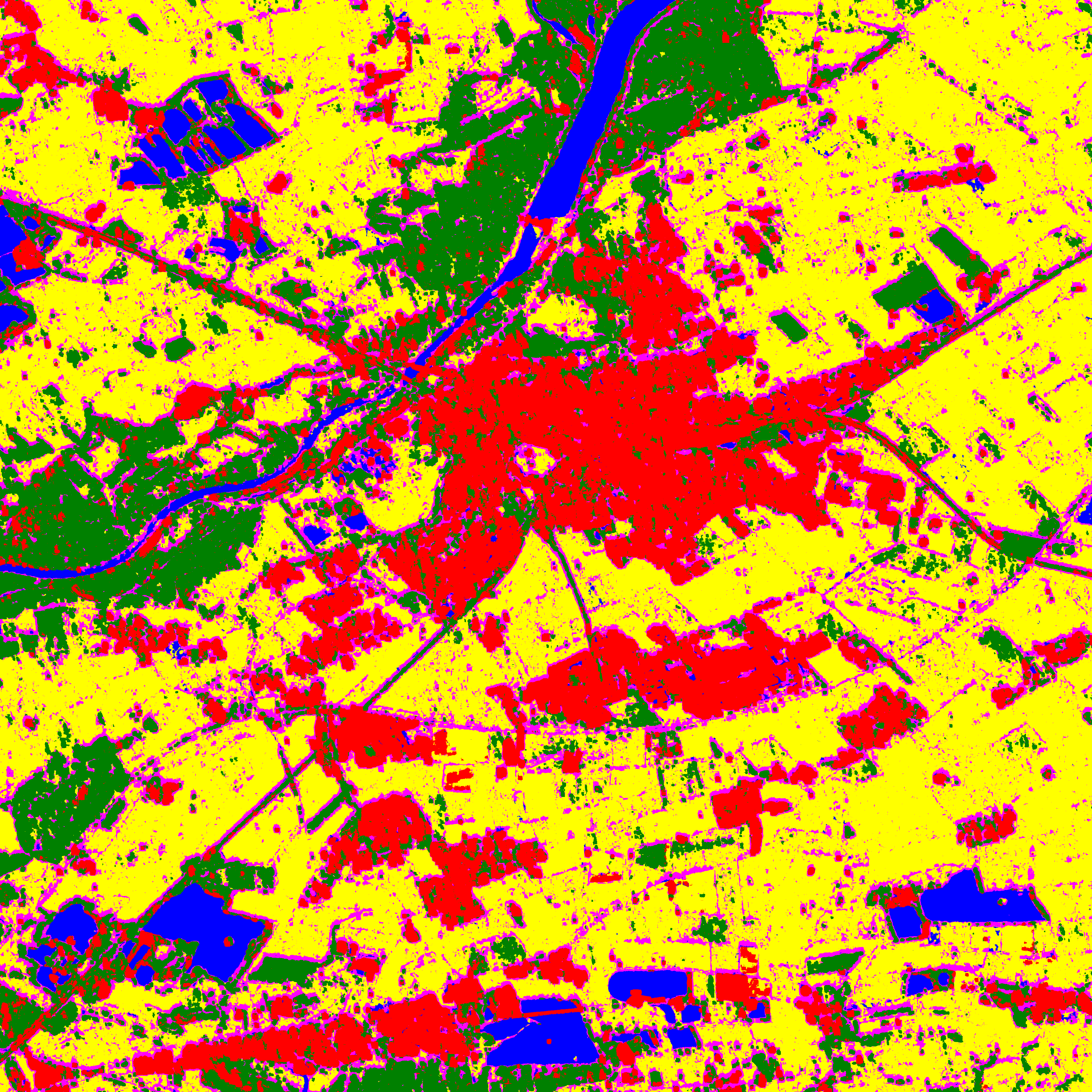}
\includegraphics[width=0.45\columnwidth]{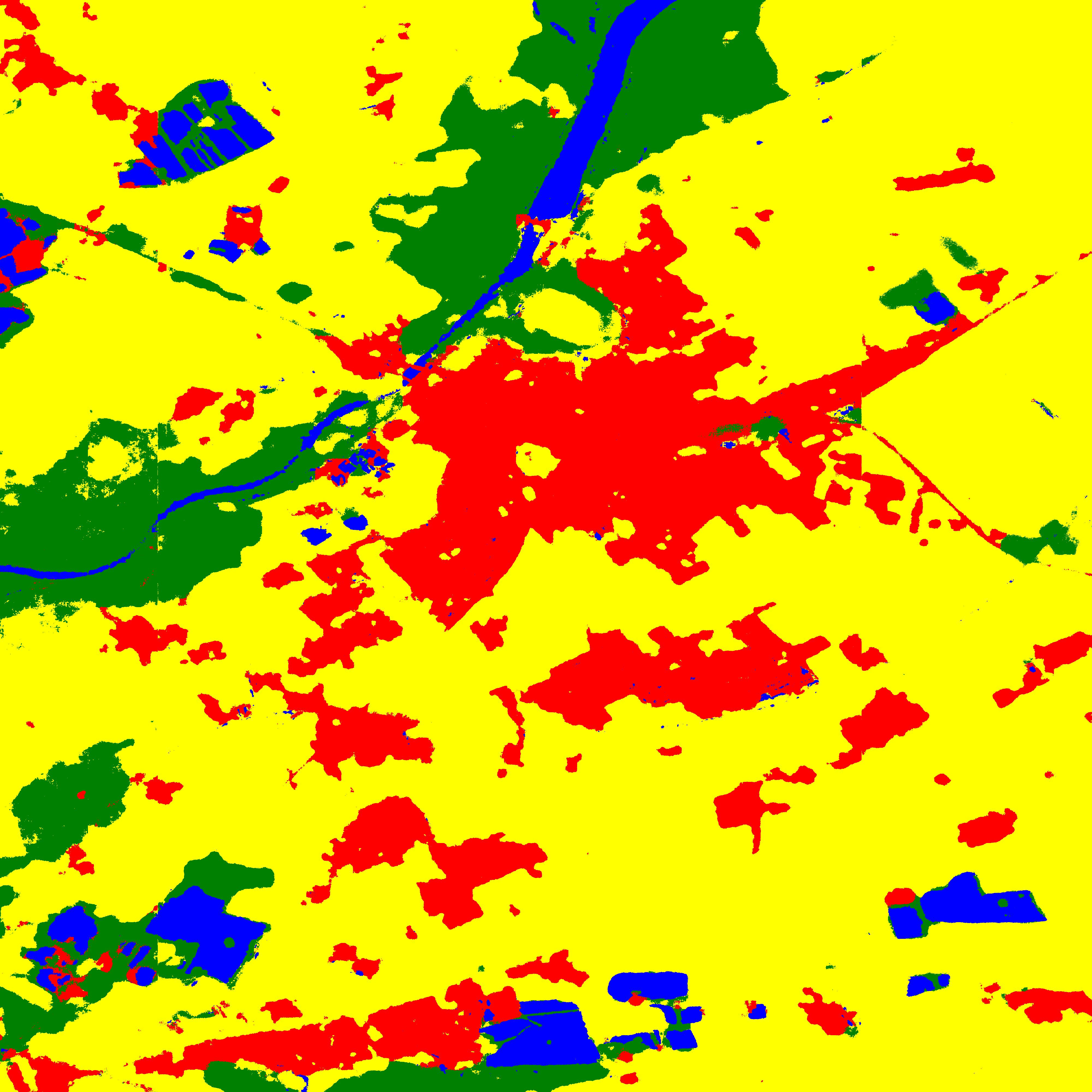}
\caption{Detail of the semantic segmentation results. First row: Pseudo-color image of the TSX data (left) and the reference map (right). Center row: Results obtained without (left) and with (right) optimization. The image shows the results of the iterative optimization approach. Results of the preselection and grouping method are similar and omitted for brevity. Bottom row: Results obtained by a RF with ten trees (left) and a Segnet (right).}
\label{fig:rfeDetail}
\end{figure}

Figure~\ref{fig:rfeDetail} shows a detailed view of the obtained semantic maps around the city of Plattling, Germany. 
The baseline solution shows a lot of small, noise-like misclassifications while the optimized RFe leads to a smoother map. One of the larger issues in both maps is that dominant line features such as the river border are classified as urban area. A possible reason is that such strong image gradients occur often around buildings. Parts of borders between agricultural fields are often misclassified as roads. Many of the roads in this dataset are between agricultural fields. Thus, the classifier has seen many examples where thin line structures between two fields do represent actual roads. Water, forest, and urban areas are very well detected and show outlines consistent with the reference data. 
The RF has mostly the same main issues as the RFe since both are based on the same class of binary features and only differ in their arrangement and optimization. 
SegNet produces a smooth map but misses on fine structures such as roads.
All methods struggle strongest with detecting roads. 
Due to the resolution of the data, most roads appear as very thin lines in the image. Thus, within a patch centered on a road, the neighborhood dominates the signal in the sense that other classes are more dominant. This states a very difficult problem which requires very specific spectral-spatial features. The amount of optimization within RFe is barely sufficient to discover these specialized features. Increasing the amount of optimization helps (i.e. accuracy of the street class raises from $38.5\%$ to more than $44\%$ if optimization techniques are applied).  
The SegNet with its receptive fields suffers from the same issues but even worse. The RF shows the same trend: The shallow trees of  {\it RF-30-8} do not allow for sufficiently specialised features while the higher trees of  {\it RF-10-30} can differentiate street patches the best.

It should be noted that the selected Deep Learning approach should only be seen as an example baseline to put the results obtained by the shallow RFe learner into perspective. 
There are certainly other architectures that would have led to different results, e.g. being focused on more fine-grained details. 
The more important part of these experiments is the comparison with the RF. 
It is based on the same node projections and is applied to the same data representation with the two main differences that the individual weak learners employ a feature hierarchy (i.e. being trees) and are combined by additive averaging, while the weak learners of the RFe (i.e. the feature groups) are "flat" and are combined by multiplication. 
This results in a model that is simpler in terms of implementation and interpretation as well as more efficient, but is able to provide results that are on par with the performance of a RF.

\section{Conclusion \& Future Work \label{sec:section5}}
In this paper we transferred the node projection of a Random Forest framework tailored towards analyzing polarimetric SAR data to the concept of Random Ferns. 
This allows to apply the proposed classifier directly to PolSAR data without the need to manually design and select features (as typically done in many other approaches for PolSAR classification). 
Furthermore, two novel optimization techniques for RFe are proposed that successfully exclude uninformative and redundant features while grouping dependent features within a single fern. 
Results show that both optimization approaches lead to similar results while being on par to the performance of a RF. 
However, being a simpler model, train and test times of the RFe are significantly shorter than for RFs. 
The comparison to a baseline deep learning model, i.e. a SegNet, shows that the obtained performance is on par with even more complicated models.

The obvious next step is to exploit the flat structure of RFe for parallelization to speed up training and application. 
Training and evaluation of each fern are independent of all other ferns and can thus run in parallel similar to parallelization approaches of RFs. 
In both cases, the base-learners are independent of each other and can thus be trained and applied in parallel. In theory, that leads to a decrease in computation time that is linear in the number of used threads. However, the random access to memory caused by the hierarchical nature of the decision trees in a RF leads to a significant amount of cache misses while RFe facilitate coalesced memory access. That is why the expected speed-up is smaller for RF than for RFe.
However, the largest potential of RFe for speed-up is by using GPUs.
RFs are not very well suited for GPUs as they are basically a large collection of if-then-else rules and do not follow the SIMD principle (i.e. different parts of the code are evaluated for different parts of the data).
In contrast to RFs, features within RFe are evaluated independently, i.e. every feature is computed for every sample. This means that the same code instructions are applied to every single sample which makes RFe a prime example for GPU processing. 
Such a GPU implementation will lead to a significant decrease in training time which would allow for stronger optimization strategies as well as larger RFe in terms of both, feature groups and number of features per group.

\ifCLASSOPTIONcaptionsoff
  \newpage
\fi

\bibliographystyle{IEEEtran}
\bibliography{./bibtex/bib/reference}

%

\begin{IEEEbiography}[{\includegraphics[width=1in,height=1.25in,clip,keepaspectratio]{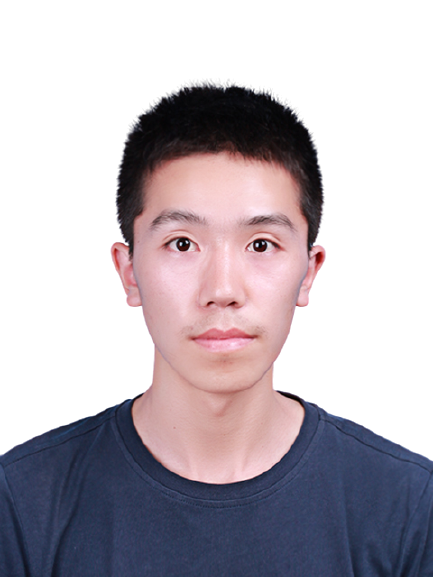}}]{Pengchao Wei}
received his B.Eng. degree in information engineering at Shanghai Jiao Tong University, Shanghai, China, in 2019, and his M.Sc. degree in computer science from Technische Universität Berlin, Germany in 2020. He is currently pursuing his M.Eng. degree in information engineering with Electronic Engineering department of Shanghai Jiao Tong university, Shanghai, China. His research interests include remote sensing image analysis, statistical machine learning, and deep learning.
\end{IEEEbiography}

\begin{IEEEbiography}[{\includegraphics[width=1in,height=1.25in,clip,keepaspectratio]{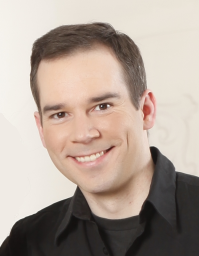}}]{Ronny Hänsch} received his Diploma in computer science and his Ph.D. degree from the Technische Universität Berlin, Germany, in 2007 and 2014, respectively. He is currently with the SAR Technology department of the German Aerospace Center (DLR) in Oberpfaffenhofen, Germany. His current research interests focus on ensemble methods and deep learning for analysis of remote sensing images in particular SAR data. He was co-chair (2017--2021) and is current chair of the IEEE Geoscience and Remote Sensing Society (GRSS) Image Analysis and Data Fusion (IADF) Technical Committee and co-chair of the International Society for Photogrammetry and Remote Sensing Working Group II/1 (Image Orientation). He serves as guest editor for IEEE JSTARS, associate editor for IEEE GRSL, and editor in chief for the GRSS e-Newsletter. 
\end{IEEEbiography}







\end{document}